\documentclass[11pt]{article}
\usepackage[preprint]{acl}
\usepackage{xcolor,colortbl}

\usepackage{booktabs}
\usepackage{times}
\usepackage{bbm}
\usepackage{latexsym}
\usepackage{amsmath}
\usepackage{amssymb}
\usepackage{placeins}
\usepackage{nameref}
\usepackage{pdflscape}
\usepackage{multirow}
\usepackage{pdfpages} 
\usepackage{makecell}
\usepackage{longtable}
\usepackage{tikz}
\usepackage{fancyvrb}
\usepackage{fvextra} 
\usepackage{hyperref}
\usepackage{adjustbox}
\usepackage{booktabs}
\newcommand{\sr}{\xi^{\mathcal{SR}}}

\newcommand{\paddedfig}[1]{%
  \vspace{0pt}%
  \hspace{8pt}%
  \includegraphics[width=2cm]{#1}%
  \hspace{8pt}%
  \vspace{0pt}%
}
\newcommand{\twoplotpaths}[3]{%
  \begin{minipage}{\textwidth}
    \includegraphics[width=0.45\textheight, trim=0 0 0 0, clip, angle=90]{#1}%
    \hspace{0.02\textwidth}
    \includegraphics[width=0.75\textwidth, trim=0 5cm 0 0, clip]{#2}
    \captionsetup[figure]{hypcap=false}
    \captionof{figure}{#3}
  \end{minipage}
}
\usepackage[T1]{fontenc}
\usepackage[utf8]{inputenc}
\usepackage{microtype}
\usepackage{longtable}
\usepackage{inconsolata}

\usepackage{graphicx}
\usepackage{xstring} 
\usepackage{xparse}
\usepackage{etoolbox}

\PassOptionsToPackage{hyphens}{url}\usepackage{hyperref}

\newcommand{\hfciteprocess}[1]{%
  \IfSubStr{#1}{https://huggingface.co/datasets/}{%
    \StrBehind{#1}{https://huggingface.co/datasets/}[\hfname]%
  }{%
    \StrBehind{#1}{https://huggingface.co/}[\hfname]%
  }%
  \href{#1}{\hfname}%
}

\newcommand{\hfcite}[2]{%
  (%
  \begingroup
    \def\hf@first{1}%
    \renewcommand*{\do}[1]{%
      \ifnum\hf@first=1
        \hfciteprocess{##1}%
        \def\hf@first{0}%
      \else
        , \hfciteprocess{##1}%
      \fi
    }%
    \docsvlist{#1}%
  \endgroup:~%
  \citealp{#2})%
}

\usepackage{enumitem}
\newcommand{\papertitle}{Compact Example-Based Explanations for Language Models}
\author{
 \textbf{Loris Schoenegger\textsuperscript{1,2}},
 \textbf{Benjamin Roth\textsuperscript{1,3}}
\\
\\
 \textsuperscript{1}Faculty of Computer Science, University of Vienna, Vienna, Austria\\
 \textsuperscript{2}UniVie Doctoral School Computer Science, University of Vienna, Vienna, Austria\\
  \textsuperscript{3}Faculty of Philological and Cultural Studies, University of Vienna, Vienna, Austria\\
\\
 \small{
   \textbf{Correspondence:} \href{mailto:loris.schoenegger@univie.ac.at}{loris.schoenegger@univie.ac.at}
 }
}

\title{\papertitle}

\newcommand{\validationSanityCheckFractionLargerLogP}{98.49\%}
\newcommand{\validationSanityCheckFractionLargerJSD}{96.28\%}
\newcommand{\validationOverallCorrelationLogP}{0.09}
\newcommand{\validationOverallCorrelationJSD}{0.07}
\newcommand{\validationLargeGainCorrelationLogP}{0.58}
\newcommand{\validationLargeGainCorrelationJSD}{0.37}

\newcommand{\validationLargeGainSettingsWhereFLDecreasesRho}{BM25 $\lambda=\{ 1.0 \}$}
\newcommand{\validationLargeGainSettingsWhereFLIncreasesRho}{LESS $\lambda=\{ 0.25,0.5,0.75,1.0 \}$}

\begin{document}
\maketitle
\begin{abstract}
Training data influence estimation methods quantify the contribution of training documents to a model’s output, making them a promising source of information for example-based explanations.
As humans cannot interpret thousands of documents, only a small subset of the training data can be presented as an explanation.
Although the choice of which documents to include directly affects explanation quality, previous evaluations of such systems have largely ignored any selection strategies.
To address this, we propose a novel \textit{selection relevance score}, a retraining-free metric that quantifies how useful a set of examples is for explaining a model's output.
We validate this score through fine-tuning experiments, confirming that it can predict whether a set of examples supports or undermines the model's predictions.
Using this metric, we further show that common selection strategies often underperform random selection. 
Motivated by this finding, we propose a strategy that balances influence and representativeness, enabling better use of selection budgets than naively selecting the highest-ranking examples.
\end{abstract}
\section{Introduction}
Training data influence estimation methods, such as \textit{influence functions} \cite{koh_understanding_2017}, estimate the contribution of individual training documents to a model's output. These estimates offer a promising source of information for creating example-based explanations for language models.
However, training data influence estimates are not directly usable as explanations. Humans cannot meaningfully process thousands of documents, nor can retrieval-augmented generation systems designed to generate natural language explanations.
To provide explanations of a human-interpretable size, existing systems rely on naive selection strategies. For example, they may choose the $k$ highest-ranking examples from the influence estimate.
This is problematic for two reasons.
First, examples in the upper tail of the influence distribution tend to be globally influential outliers. These outliers are not necessarily the most relevant for the test instance \cite{barshan_relatif_2020}.
Second, the highest-ranking training examples often contain redundant information \cite{bhatt_divine_2021}. Strictly selecting only the most influential documents can yield diminishing returns, as these examples may offer little new information to users.
Similarly, when selecting documents as a basis for generating natural language explanations, choosing outlier or redundant examples can undermine explanation faithfulness.
\begin{figure}[ht]
\centering
  \includegraphics[width=1\columnwidth, trim=0 0 0 0, clip]{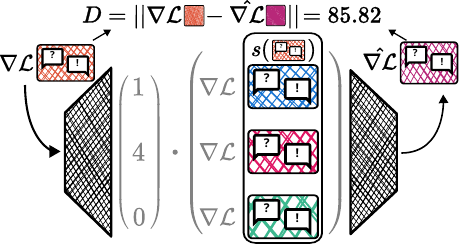}
  \caption {\textbf{Simplified scoring setup}. We score example-based explanations based on how well the test instance's loss gradient (left) can be reconstructed (right) as a linear combination of the selected examples' gradients.}\label{fig:eval_setup}
\end{figure}

Although example selection directly affects explanation quality, existing evaluations focus on the correctness of raw estimates or on testing the usefulness of the full explanation system with users. This is problematic as it can obscure flaws in the selection logic: a system may appear helpful even if the selected examples are not truly informative. 

In this paper, we propose an evaluation score that \textbf{specifically targets the selection logic}.
For this, we frame example-based explanation as a reconstruction task (Figure~\ref{fig:eval_setup}): specifically, we propose the measure of \textit{selection relevance}, which quantifies how well the gradients of the selected examples reconstruct the gradient of the test instance they are meant to explain.
Intuitively, selections truly relevant to the test instance should yield lower reconstruction errors than sets of globally influential outliers or redundant examples.
Notably, this score considers the relevance of the selected examples \textbf{in combination}. In contrast, existing approaches either assess individual examples in isolation or evaluate groups only in costly retraining-based experiments.
Our setup is task- and estimation-method agnostic. It can also be easily adapted to additional explanation-specific selection constraints.

We validate our proposed selection relevance score by comparing it to an alternative notion of relevance derived from fine-tuning experiments:
two fine-tuning-based metrics capture how training on the selection affects the original prediction.
As neither notion of relevance provides a definitive ground truth, we assess their alignment through correlation analysis.
We further demonstrate the utility of our score by evaluating various combinations of influence estimation methods and selection strategies.
We show that common selection strategies often underperform random baselines and introduce a new strategy that more effectively leverages human-interpretable selection budgets than simply choosing the most influential examples.
Our main contributions are as follows:\footnote{We release our code at \href{https://doi.org/10.5281/zenodo.19483839}{doi.org/10.5281/zenodo.19483839}.}

\begin{enumerate}[nosep, topsep=0pt, label=(\arabic*), left=0pt]
    \item We propose the \textit{selection relevance score}, a retraining-free selection quality score that evaluates the utility of a set of training documents for providing example-based explanations.
    \item We benchmark 3 influence estimation and 4 selection strategies, providing insights into why naive selection strategies are ineffective if paired with popular estimation methods. 
    \item Using our score, we introduce a new strategy that better leverages small selection budgets.
\end{enumerate}
\section{Background and Related Work}\label{rw}
\paragraph{Evaluating influence estimates.}
For influence estimates to be a useful source of information for explanations, they must accurately represent how individual training examples affect the model's behavior during training.
Existing work on evaluating the \textit{correctness} of influence rankings has largely followed two approaches: testing how well rankings correlate with results from leave-one-out (LOO) retraining \cite{park_trak_2023,bae_if_2022,basu_influence_2020,k_revisiting_2021,choe_what_2024}, or assessing performance in downstream data discovery tasks, such as data pruning \cite{koh_understanding_2017}.
However, for our task, assessing 
correctness 
is insufficient, we target the \textit{selection mechanism}, which is difficult to isolate from other components in downstream task evaluations.
\paragraph{Evaluating explanations.}
We observe that existing example-based explanation methods share a common motivation: 
they assume that an explanation can only faithfully represent model behavior if the selected examples are sufficiently relevant to the output being explained. For example, some approaches define relevance in terms of embedding similarity between the test instance and explanation elements \cite[e.g.,][]{zhang_sample_2021, nematov_susceptibility_2024}, or by checking if the selected instances share the label of the test instance \cite{hanawa_evaluation_2020}. Still others verify whether training only on the selected instances can reproduce the outputs obtained from training on the full dataset \cite[e.g.,][]{gu_iaeval_2023}, which also requires sufficient relevance to test instances.
However, these measures either operate in embedding space (while ranking typically occurs in gradient space), rely on class labels (making them unsuitable for text generation tasks), or require retraining (which is intractable for LLMs). Our score avoids all of these limitations.
\paragraph{Locally vs.~globally influential training data.}
\citet{barshan_relatif_2020} observe that influence functions tend to assign high influence to a small, consistent set of training examples across test instances, which frequently consists of outliers or mislabeled examples.
Similarly, \citet{hammoudeh_identifying_2022} argue that popular influence estimation methods underestimate the influence of highly influential
instances because confidently predicted examples have small gradient magnitudes.
Additionally, \citet{nematov_susceptibility_2024} find that examples that consistently rank highly across many test instances tend to have high loss.
Based on the idea that showing outliers to users does not constitute good example-based explanations \cite{barshan_relatif_2020}, some works have attempted to adjust for the influence of globally influential examples in new influence estimation methods \cite{barshan_relatif_2020,nematov_aide_2024,hammoudeh_identifying_2022}. 
However, none explicitly account for the difference between globally and locally relevant examples in their evaluations. In contrast, our work directly evaluates the \textit{local relevance} of \textit{selections}.
\paragraph{Reducing redundancy within local explanations.}
Observing that the most influential examples often form clusters in feature space, both \citet{bhatt_divine_2021} and \citet{nematov_aide_2024} consider example diversity alongside influence.
Both 
utilize objectives of the form $\max_{S \in \mathcal{D},\, |S| = k} \; I(S) + D(S)$,
where $D$ quantifies diversity and $I$ measures influence.
However, this strategy can favor outliers (as \citeauthor{bhatt_divine_2021} note), which we argue is problematic given the 
behavior discussed above.
In this work, we instead aim to increase \textit{representativeness}. 
\paragraph{Prediction-constrained influence.}\label{pred_constr_inf}
Influence functions sometimes poorly predict re-training effects for deep models not trained to convergence \cite{bae_if_2022,basu_influence_2020,grosse_studying_2023}.
\citet{bae_if_2022} partially attribute this to the fact that the re-trained model's parameters and predictions are not guaranteed to remain sufficiently close to those of the original model after LOO retraining. They propose an alternative measure to LOO-influence, which measures the effect of removing examples while penalizing deviations from the model's original predictions.% 
 They find that influence functions correlate more strongly with this \textit{prediction-constrained} measure than with the LOO ground truth, and argue that it is more useful in practice than measuring LOO effects directly.
Prediction-constrained influence appears compatible with our task: local explanations are meaningful only as long as the model is not altered so much that it deviates from the original prediction \cite{ribeiro_why_2016,guidotti_survey_2018}.
However, there is no guarantee that this is considered during example selection: the set may instead support an entirely different decision.
We therefore also evaluate how our score aligns with this alternative view of relevance (Section \ref{validation_score}).
\section{Methodology}\label{methodology}
We study the task of selecting training examples for example-based explanation, where we aim to explain a model's prediction $(\hat{y} = f_\theta(x))$ through a small set of training examples.
In this section, we introduce a novel evaluation setup for this task. We then validate it in Section \ref{results} and also use it to benchmark various combinations of influence estimation methods and selection strategies.

The criteria for a good example-based explanation are application-specific \cite{barshan_relatif_2020}. In this paper, we focus on explanations that present \textit{training examples supporting the prediction of interest} \cite[e.g.,][]{barshan_relatif_2020,hanawa_evaluation_2020,yeh_representer_2018}, rather than alternative setups such as selecting counterfactual examples that would change the model's prediction.
Moreover, we assume that attribution targets prediction-constrained influence (see Section \ref{rw}) at the final model state and that estimates are computed with respect to a single test instance.

The training data influence estimation methods we use (Section \ref{experimental_setup}) produce a ranking over the entire training dataset ($\phi(f, x, \hat{y}) \in \mathbb{R}^{||D||}$). However, we evaluate selections $S$ of human-interpretable sizes $k = \{1,5,10,25\}$.  
In this section, we introduce a \textit{selection relevance score} $\sr$.
For illustration, we consider a simple selection method $s$ that chooses the $k$ highest-ranked training examples according to some influence estimate.

\subsection{Evaluation Setup}
Our score quantifies how \textit{relevant} the chosen training examples are for explaining the model output $\hat{y}$ of interest.
We specifically consider relevance at the selection level: 
Given the small selection budget, we argue that the most influential examples from the training data are not necessarily the most useful from the user's perspective. 
Providing users with a document that is highly influential but largely redundant with other examples is less likely to improve their understanding of the model.
The selection must represent a sufficiently diverse set of training behaviors to adequately explain the output.
We operationalize this notion of relevance as a gradient reconstruction task:
\paragraph{Encoding.}
Our scoring approach uses model loss gradients, in line with popular influence estimators that also rely exclusively on gradients.
We aim to reconstruct the loss gradient of a given test instance, $\nabla \mathcal{L}'$, using a linear combination, $\hat{\nabla \mathcal{L}}' = A t$,
where $t \in \mathbb{R}^{k}$ denotes a set of learned coefficients to be introduced in Section \ref{scoring_models}, and $A$ is a matrix of gradients of the $k$ selected training examples:
$A = [\, \nabla \mathcal{L}_1 \ \nabla \mathcal{L}_2 \ \cdots \ \nabla \mathcal{L}_k \,], \;
\nabla \mathcal{L}_i \in \mathbb{R}^d$. 
\paragraph{Selection relevance score.}
We interpret the reconstruction error as a measure of the relevance of the selected examples for a given test instance.
Let $\mathcal{D} = \{\mathbf{y}^{(n)}\}_{n=1}^N \subset \Sigma^*$ be the dataset used in training, drawn from the ground-truth distribution $p_{\theta^*}$:
$
\mathbf{y}^{(n)} \sim p_{\theta^*}, \quad n = 1, \dots, N.
$
Let $Y$ be a random variable representing training examples uniformly sampled from $\mathcal{D}$.
We define a random vector $G: \Omega \to \mathbb{R}^d$ representing gradients of the model's loss function 
$
G = \nabla_\theta \mathcal{L}(Y, \hat{Y};\theta)
$.
Each realization \(G(\omega_0)\), \(\omega_0 \in \Omega\) corresponds to the gradient of the loss with respect to the model parameters for a single training example. 
A single \(G(\omega_0)\) can be reconstructed by $
\hat{G}(\omega_0) = A t_0 $.
The reconstruction error is then defined as $D(\omega_0) \in \mathbb{R}^d$:
$
D(\omega_0) = G(\omega_0) - \hat{G}(\omega_0) = G(\omega_0) - A t_0
$.
Extending this reconstruction to all realizations of $G(\omega)$, $\omega \in \Omega$, we approximate $G(\omega), \omega \in \Omega$ as
$
\hat{G}(\omega) = A t_\omega, \quad 
D(\omega) = G(\omega) - \hat{G}(\omega)
$. The coefficients \(t_\omega\) are optimized individually for each \(G(\omega)\), \(\omega \in \Omega\).
We define the \textit{selection quality score} $\sr$ as the ratio of the expected squared gradient norm to expected squared reconstruction error:
\begin{align}
\sr &=
\frac{\mathbb{E}_\omega [ \| G(\omega) \|^2 ]}
     {\mathbb{E}_\omega [ \| D(\omega) \|^2 ]}\\
&= \frac{\mathbb{E}_\omega [ \| G(\omega) \|^2 ]}
       {\mathbb{E}_\omega [ \| G(\omega) - A t_\omega \|^2 ]}\notag
\end{align}
\paragraph{Interpretation.}
We report this score in decibels ($10 \log_{10} \sr$) throughout to ease visualization.
Our score $\sr$ quantifies how well the chosen training examples can reconstruct the gradient of a test instance, capturing their informativeness. Values below $0\ \mathrm{dB}$ (i.e., $\sr < 1$ in absolute scale) indicate worse approximation than the trivial baseline $||G - \vec{0}|| $,
meaning the selected examples fail to convey relevant information. Values above $0\ \mathrm{dB}$ ($\sr > 1$) suggest 
non-redundant information.

\subsection{Scoring Models}\label{scoring_models}
In practice, multiple $t$ can minimize the approximation error, 
among them the least squares solution
$
t^* = (A^\top A)^{-1} A^\top \nabla \mathcal{L}' 
$.
However, we impose two additional constraints on the solution to ensure consistency with the assumptions underlying the explanation process outlined at the beginning of Section \ref{methodology}:
First, we impose a non-negativity constraint on the coefficients $t$.
This ensures the weights are non-negative, preventing cancellations between irrelevant examples. More generally, we prefer that all examples either provide evidence for or evidence against the prediction (i.e., all either increase or decrease loss, or have a strong or weak effect on $\theta$).
Second, by enforcing $\sum t = 1$, we effectively normalize the importance scores across selections. This allows users to interpret $t$ as a vector of relative importance within the selected set.
To account for both constraints while preserving an analytical solution, we first compute an unconstrained least-squares solution, and then project it onto the unit simplex to obtain a normalized, non-negative vector. 
We chose this implementation because directly optimizing the constrained objective in a training-based setup would require monitoring convergence and tuning optimizer hyperparameters. 
We consider alternative constraints in Appendix \ref{appendix:alt_constr}.
\subsection{Fine-Tuning-Based Validation Experiment}\label{validation_score}
As discussed in Section \ref{pred_constr_inf}, local explanations may no longer hold if the model is altered in a way that causes it to produce substantially different outputs.
However, there is no guarantee that this aspect is considered during example selection: the selected set could inadvertently support a different decision.
To assess whether selections with high selection relevance scores also sufficiently support the original predictions, we report Spearman correlations with two fine-tuning-based metrics:
\paragraph{Prediction support $\xi^+$.} For each test instance-, model- and estimator pairing, we train for one step on the documents in the selection (LR=1e-5; one batch) and measure the impact on the original prediction.
Specifically, we calculate the likelihood of the originally generated output $\log p(y|x; \cdot)$ for a model that was trained for one additional step on the selected examples $S$ ($\log p(y|x; \theta_{+S})$), and for a model trained on a random subset $R$ of the training data ($\log p(y|x; \theta_{+R})$; $|R| = |S|$).
The intuition is that fine-tuning on truly informative examples should increase the likelihood of the original output more than training on a random set.
The following score indicates whether $S$ supports the original output more than a random selection $R$:
\begin{align}
\xi^+(S) &=  \log p(y \mid x; \theta_{+S})\\
        &\quad - \log p(y \mid x; \theta_{+R})\notag
\end{align}

\paragraph{Prediction shift $\xi^{JSD}$.}
Additionally, we measure Jensen-Shannon divergences to capture whether $S$ induces a larger shift in the model's full predicted distribution $p(y \mid x; \cdot)$ than a random subset would:
\begin{align}
\resizebox{0.8\linewidth}{!}{$
\xi^{JSD}(S) = 
\frac{
    \mathrm{JSD}\big(p(y \mid x; \theta),\, p(y \mid x; \theta_{+S})\big)
}{
    \mathrm{JSD}\big(p(y \mid x; \theta),\, p(y \mid x; \theta_{+R})\big)
}
$}
\end{align}
\begin{figure*}[!ht]
\centering
  \includegraphics[width=1\textwidth, trim=0 0 0 0, clip]{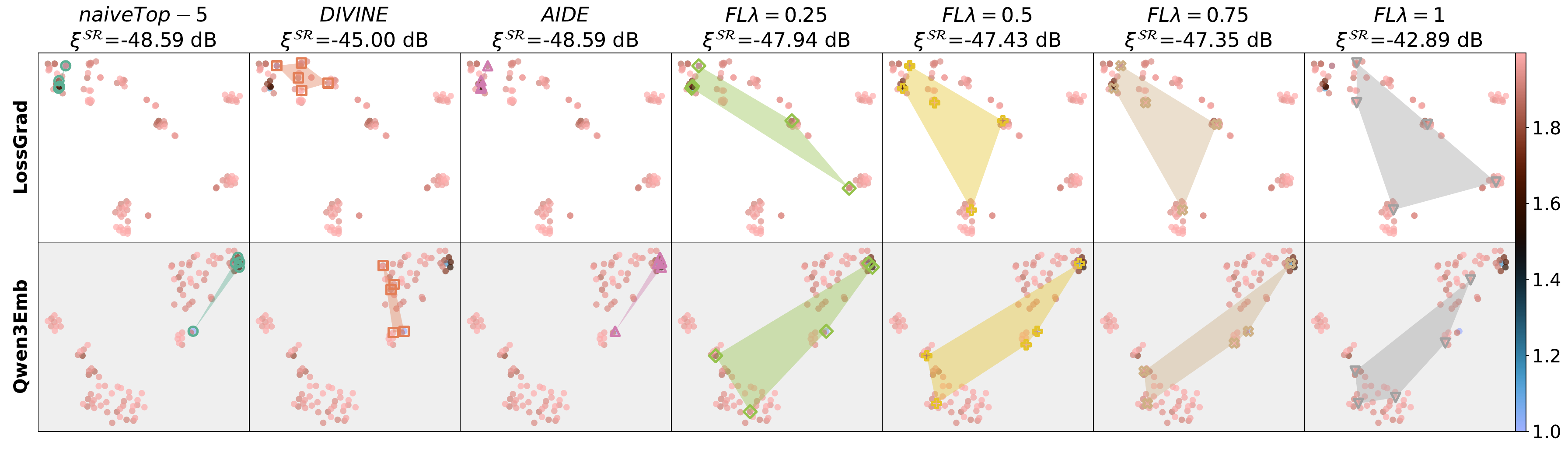}
  \caption {\textbf{Case Study}.  Each column highlights $k\!=\!5$ documents selected by a selection method from the $m\!=\!100$ \textit{most influential} documents identified by DataInf. Shaded regions indicate coverage. t-SNE visualizations of the 100 most-influential examples in gradient and embedding spaces (\href{https://huggingface.co/Qwen/Qwen3-Embedding-0.6B}{Qwen/Qwen3-Embedding-0.6B}: \citealp{zhang_qwen3_2025}).
  Color indicates selection cost (higher influence corresponds to lower cost). Our selections (FL) improve coverage.
  }\label{fig:case_study_1_tsne}
\end{figure*}
\subsection{Experimental Setup}\label{experimental_setup}
We use DataInf \cite[approximates influence functions;][]{kwon_datainf_2024}, and LESS \cite[gradient similarity;][]{xia_less_2024} to obtain influence estimates.
Both methods restrict influence estimation to the LoRA layers. 
Additionally, we include a BM25 baseline that retrieves training examples based on their token overlap with the test instance. We discuss parameter choices in detail in Appendix \ref{appendix:tda}.

\paragraph{Naive selection.}
Given an influence estimate $\phi(f, x, \hat{y}) \in \mathbb{R}^{||D||}$, we select $k$ documents according to the following strategies: lowest influence scores \cite[\textit{most helpful:}][]{koh_understanding_2017}, largest influence scores (\textit{most harmful}), largest absolute scores (\textit{most influential}), and lowest absolute scores (\textit{least influential}). Additionally, we report the mean performance of 5 random selections for each $k$.
\paragraph{Coverage-aware selection.} 
To systematically evaluate our scoring method, we implement a selection strategy that optimizes for example coverage. 
Our goal is to select a more representative, less redundant set of training examples than the naive selection method. We therefore treat selection as a \textit{facility location problem} \cite[see e.g.,][]{Krause_Golovin_2014}. 
Facility location functions are typically defined as
$F(S) = \sum_{i \in V} \max_{j \in S} \operatorname{sim}_{ij}$,
where $S \subseteq V$ is the selected subset. 
The marginal gain of adding an element $j \in V$ to the current set $S$ is
$\Delta(j \mid S) = F(S \cup \{j\}) - F(S) $.
We extend this formulation to incorporate influence scores:
\begin{align}
\resizebox{0.6\columnwidth}{!}{$
\Delta_\lambda(j \mid S) 
= \frac{(\Delta(j \mid S) + 1)^\lambda}{c_j^{\,1-\lambda}},
$}
\end{align}
where $c_j$ is a normalized cost score based on the training data influence of element $j$.
Setting $\lambda = 1$ performs a purely coverage-based selection, while $\lambda = 0$ is identical to the naive selection by influence scores.
We implement selection via a custom optimizer in the \verb|apricot| library \cite{schreiber_apricot_2020}: we greedily select the example with the highest $\Delta_\lambda(j \mid S)$ among the top-100 examples in the naive ranking until the budget is exhausted.
\paragraph{Diversity-based selection.}
We reimplement \mbox{DIVINE} \cite{bhatt_divine_2021} as its code is not publicly available, and remove elements from AIDE \cite{nematov_aide_2024} that are only applicable to classification tasks. See Appendix \ref{appendix:divine_aide} for details.
\paragraph{Datasets and models.}
We include models from three families: Olmo2 \hfcite{https://huggingface.co/allenai/OLMo-2-0425-1B}{olmo_2_2025}, Llama~3.2 \hfcite{https://huggingface.co/meta-llama/Llama-3.2-1B}{dubey_llama_2024}, and Qwen~2.5 \hfcite{https://huggingface.co/Qwen/Qwen2.5-0.5B}{yang_qwen25_2024}.
We fine-tune for one epoch using LoRA \cite{hu_lora_2021} on the full \textit{Tülu3} \hfcite{https://huggingface.co/datasets/allenai/tulu-3-sft-olmo-2-mixture-0225}{lambert_tulu_2025} instruction-fine tuning dataset, see Appendix \ref{appendix:finetuning} for hyperparameters and benchmark results. 
For training data attribution, we randomly sample 10\% (86.6k) of the examples as the set to attribute to, and a disjoint set of 1,000 test instances to explain.

\begin{figure}[htb]  
\centering
  \includegraphics[width=1\columnwidth, trim=0.05cm 4.75cm 0.2cm 0.025cm, clip]{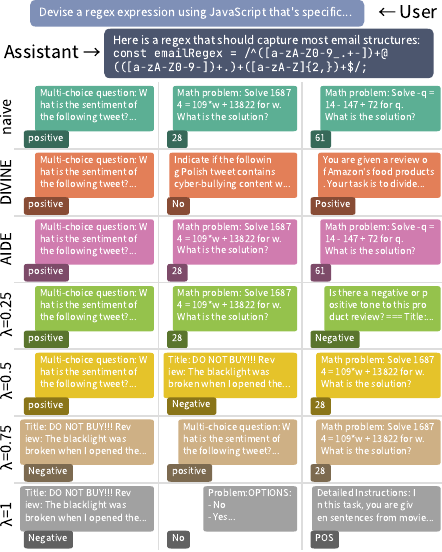}
  \caption {\textbf{Case Study}. First 3 documents per selection ($k$=5, DataInf, most inf.). Full figure in Appendix \ref{appendix:case_study}.}\label{fig:onecol_case_study}
\end{figure}
\section{Results}\label{results}
We first present a \textbf{case study} to illustrate how our proposed selection relevance score relates to coverage and redundancy 
in gradient- and embedding spaces.
In particular, we aim to provide intuition about how different 
choices of $\lambda$ affect coverage in facility-location-based selections, to aid in the interpretation of subsequent quantitative evaluation.
Note that we do not aim to demonstrate the utility of specific selection methods for producing user-facing explanations here, nor to demonstrate the utility of example-based explanations in general. Both would require dedicated user studies, which are beyond the scope of this work.

We then \textbf{benchmark} estimation methods (Section \ref{sel_qual_scoring}) and \textbf{validate} our scoring setup using two complementary approaches: an experiment with the facility-location-based method to show that improved coverage increases scores (Figure \ref{fig:relative_improvement}), and a validation experiment assessing correlation with two training-based scores (Section \ref{validaiton_experiment_results}).
\subsection{Case Study}\label{case_study}
We present selections for a single test instance derived from a DataInf influence estimate for the Olmo2 model.
In Figure \ref{fig:case_study_1_tsne}, we plot t-SNE visualizations of the 100 most-influential examples and highlight the selections for a budget of $k$=5. 
Color indicates influence scores rescaled to selection costs in the range $[1, m]$ using min-max normalization, where higher influence corresponds to lower cost.

Facility location-based strategies 
 improve coverage over naive selection in both gradient- and embedding spaces. Consistent with the intended behavior of our scoring setup, these strategies also achieve higher selection relevance $\sr$.
We additionally provide the first three examples for each method in text form in Figure \ref{fig:onecol_case_study} (see Appendix \ref{appendix:case_study} for the full selection).
The naive selection and AIDE include examples with highly similar prompts.
Only two of these redundant examples are included when $\lambda=0.25$, and only one appears when $\lambda = \{0.5, 0.75\}$.
 The purely coverage-based selection $\lambda=1$ retains none of the examples selected by the naive selection.
For this particular test instance, AIDE produces the same selection as the naive strategy. DIVINE appears to prioritize diversity more aggressively than AIDE in our setup. 
Note that both require manual hyperparameter selection, which makes their behavior difficult to interpret on a per-example basis (Appendix \ref{appendix:divine_aide}).

\begin{table*}[htbp]
\centering
\scriptsize
\setlength{\tabcolsep}{0.5pt} 
\begin{minipage}[t]{0.32\textwidth}
    \vspace{0pt}
    \centering
    \begin{tabular}{l|l|l}
\toprule
\midrule
most inf. & {\cellcolor[HTML]{FFF7FB}} \color[HTML]{000000} -22.87 dB \\
most harmful & {\cellcolor[HTML]{FDF5FA}} \color[HTML]{000000} -22.50 dB \\
most helpful & {\cellcolor[HTML]{F6EFF7}} \color[HTML]{000000} -21.52 dB \\
 AIDE & {\cellcolor[HTML]{F1EBF4}} \color[HTML]{000000} -20.70 dB \\
 FL most inf. $\lambda=.25$ & {\cellcolor[HTML]{F1EBF4}} \color[HTML]{000000} -20.65 dB \\
 FL most harmful $\lambda=.25$ & {\cellcolor[HTML]{EEE9F3}} \color[HTML]{000000} -20.34 dB \\
 FL most inf. $\lambda=.5$ & {\cellcolor[HTML]{EEE8F3}} \color[HTML]{000000} -20.26 dB \\
 FL most inf. $\lambda=.75$ & {\cellcolor[HTML]{ECE7F2}} \color[HTML]{000000} -19.99 dB \\
 FL most harmful $\lambda=.5$ & {\cellcolor[HTML]{ECE7F2}} \color[HTML]{000000} -19.94 dB \\
 FL most helpful $\lambda=.25$ & {\cellcolor[HTML]{E9E5F1}} \color[HTML]{000000} -19.75 dB \\
 FL most harmful $\lambda=.75$ & {\cellcolor[HTML]{E8E4F0}} \color[HTML]{000000} -19.63 dB \\
 FL most inf. $\lambda=1$ & {\cellcolor[HTML]{E7E3F0}} \color[HTML]{000000} -19.54 dB \\
 FL most helpful $\lambda=.5$ & {\cellcolor[HTML]{E7E3F0}} \color[HTML]{000000} -19.49 dB \\
 FL most helpful $\lambda=.75$ & {\cellcolor[HTML]{E3E0EE}} \color[HTML]{000000} -19.13 dB \\
 FL most harmful $\lambda=1$ & {\cellcolor[HTML]{E1DFED}} \color[HTML]{000000} -18.91 dB \\
 DIVINE most helpful & {\cellcolor[HTML]{DFDDEC}} \color[HTML]{000000} -18.69 dB \\
 FL most helpful $\lambda=1$ & {\cellcolor[HTML]{DAD9EA}} \color[HTML]{000000} -18.23 dB \\
 DIVINE most inf. & {\cellcolor[HTML]{D2D2E7}} \color[HTML]{000000} -17.31 dB \\
 DIVINE most harmful & {\cellcolor[HTML]{CED0E6}} \color[HTML]{000000} -17.09 dB \\
random & {\cellcolor[HTML]{045585}} \color[HTML]{F1F1F1} -2.57 dB \\
 DIVINE least inf. & {\cellcolor[HTML]{023858}} \color[HTML]{F1F1F1} -0.14 dB \\
least inf. & {\cellcolor[HTML]{023858}} \color[HTML]{F1F1F1} -0.14 dB \\
 FL least inf. $\lambda=.75$ & {\cellcolor[HTML]{023858}} \color[HTML]{F1F1F1} -0.13 dB \\
 FL least inf. $\lambda=.5$ & {\cellcolor[HTML]{023858}} \color[HTML]{F1F1F1} -0.13 dB \\
 FL least inf. $\lambda=.25$ & {\cellcolor[HTML]{023858}} \color[HTML]{F1F1F1} -0.13 dB \\
 FL least inf. $\lambda=1$ & {\cellcolor[HTML]{023858}} \color[HTML]{F1F1F1} -0.13 dB \\
\bottomrule
\end{tabular}

    \vspace{-2.5mm}
    \caption*{DataInfEstimator}
\end{minipage}
\hfill
\begin{minipage}[t]{0.32\textwidth}
    \vspace{0pt}
    \centering
    \begin{tabular}{l|l|l}
\toprule
\midrule
most inf. & {\cellcolor[HTML]{FFF7FB}} \color[HTML]{000000} -22.28 dB \\
most helpful & {\cellcolor[HTML]{FCF4FA}} \color[HTML]{000000} -21.81 dB \\
most harmful & {\cellcolor[HTML]{FCF4FA}} \color[HTML]{000000} -21.79 dB \\
 FL most inf. $\lambda=.25$ & {\cellcolor[HTML]{F4EDF6}} \color[HTML]{000000} -20.56 dB \\
 AIDE & {\cellcolor[HTML]{F2ECF5}} \color[HTML]{000000} -20.39 dB \\
 FL most inf. $\lambda=.5$ & {\cellcolor[HTML]{F1EBF5}} \color[HTML]{000000} -20.21 dB \\
 FL most harmful $\lambda=.25$ & {\cellcolor[HTML]{F1EBF4}} \color[HTML]{000000} -20.16 dB \\
 FL most helpful $\lambda=.25$ & {\cellcolor[HTML]{F0EAF4}} \color[HTML]{000000} -20.09 dB \\
 FL most helpful $\lambda=.5$ & {\cellcolor[HTML]{EEE9F3}} \color[HTML]{000000} -19.78 dB \\
 FL most inf. $\lambda=.75$ & {\cellcolor[HTML]{EDE8F3}} \color[HTML]{000000} -19.64 dB \\
 DIVINE most inf. & {\cellcolor[HTML]{EDE8F3}} \color[HTML]{000000} -19.63 dB \\
 DIVINE most harmful & {\cellcolor[HTML]{EBE6F2}} \color[HTML]{000000} -19.38 dB \\
 FL most harmful $\lambda=.5$ & {\cellcolor[HTML]{EAE6F1}} \color[HTML]{000000} -19.30 dB \\
 FL most inf. $\lambda=1$ & {\cellcolor[HTML]{EAE6F1}} \color[HTML]{000000} -19.28 dB \\
 FL most harmful $\lambda=.75$ & {\cellcolor[HTML]{E9E5F1}} \color[HTML]{000000} -19.25 dB \\
 FL most helpful $\lambda=.75$ & {\cellcolor[HTML]{E9E5F1}} \color[HTML]{000000} -19.22 dB \\
 FL most harmful $\lambda=1$ & {\cellcolor[HTML]{E7E3F0}} \color[HTML]{000000} -19.07 dB \\
 DIVINE most helpful & {\cellcolor[HTML]{E5E1EF}} \color[HTML]{000000} -18.80 dB \\
 FL most helpful $\lambda=1$ & {\cellcolor[HTML]{E1DFED}} \color[HTML]{000000} -18.42 dB \\
random & {\cellcolor[HTML]{045687}} \color[HTML]{F1F1F1} -2.57 dB \\
 DIVINE least inf. & {\cellcolor[HTML]{023858}} \color[HTML]{F1F1F1} -0.15 dB \\
least inf. & {\cellcolor[HTML]{023858}} \color[HTML]{F1F1F1} -0.14 dB \\
 FL least inf. $\lambda=1$ & {\cellcolor[HTML]{023858}} \color[HTML]{F1F1F1} -0.13 dB \\
 FL least inf. $\lambda=.75$ & {\cellcolor[HTML]{023858}} \color[HTML]{F1F1F1} -0.13 dB \\
 FL least inf. $\lambda=.5$ & {\cellcolor[HTML]{023858}} \color[HTML]{F1F1F1} -0.13 dB \\
 FL least inf. $\lambda=.25$ & {\cellcolor[HTML]{023858}} \color[HTML]{F1F1F1} -0.13 dB \\
\bottomrule
\end{tabular}

    \vspace{-2.5mm}
    \caption*{LESSEstimator}
\end{minipage}
\hfill
\begin{minipage}[t]{0.32\textwidth}
    \vspace{0pt}
    \centering 
    \begin{tabular}{l|l|l}
\toprule
\midrule
random & {\cellcolor[HTML]{FFF7FB}} \color[HTML]{000000} -2.57 dB \\
 DIVINE least inf. & {\cellcolor[HTML]{FFF7FB}} \color[HTML]{000000} -2.51 dB \\
 FL least inf. $\lambda=1$ & {\cellcolor[HTML]{FEF6FB}} \color[HTML]{000000} -2.41 dB \\
 FL least inf. $\lambda=.75$ & {\cellcolor[HTML]{FBF3F9}} \color[HTML]{000000} -1.57 dB \\
 FL least inf. $\lambda=.5$ & {\cellcolor[HTML]{F9F2F8}} \color[HTML]{000000} -1.25 dB \\
 FL least inf. $\lambda=.25$ & {\cellcolor[HTML]{F8F1F8}} \color[HTML]{000000} -1.09 dB \\
least inf. & {\cellcolor[HTML]{F8F1F8}} \color[HTML]{000000} -1.01 dB \\
 FL most inf. $\lambda=1$ & {\cellcolor[HTML]{589EC8}} \color[HTML]{F1F1F1} 15.50 dB \\
 AIDE & {\cellcolor[HTML]{056FAE}} \color[HTML]{F1F1F1} 22.08 dB \\
most inf. & {\cellcolor[HTML]{056EAD}} \color[HTML]{F1F1F1} 22.24 dB \\
 DIVINE most inf. & {\cellcolor[HTML]{05659F}} \color[HTML]{F1F1F1} 23.77 dB \\
 FL most inf. $\lambda=.25$ & {\cellcolor[HTML]{023C5F}} \color[HTML]{F1F1F1} 29.44 dB \\
 FL most inf. $\lambda=.75$ & {\cellcolor[HTML]{023858}} \color[HTML]{F1F1F1} 29.99 dB \\
 FL most inf. $\lambda=.5$ & {\cellcolor[HTML]{023858}} \color[HTML]{F1F1F1} 30.00 dB \\
\bottomrule
\end{tabular}

    \vspace{-2.5mm}
    \caption*{BM25Estimator}
    \vspace{0.1cm}
    \begin{adjustbox}{scale=0.85,center}
    \scriptsize
    \setlength{\tabcolsep}{3pt} 
    \begin{tabular}{|p{0.8cm}|p{0.9cm}|p{2.5cm}|}
    \hline
    \textbf{Label} & \textbf{Selection} & \textbf{Effect}\\
    \hline
    Most helpful & Most negative & Loss likely increases if selection removed. \\
    \hline 
    Most harmful & Most positive & Loss likely decreases if selection removed. \\
    \hline
    Most infl. & Large\newline  absolute & Strongest impact on parameter update. \\
    \hline
    Least infl.  & Small\newline  absolute & Weakest impact on parameter update. \\
    \hline
    \end{tabular}
    \end{adjustbox}
    \vspace{-2.5mm}
    \caption*{Naive selection logics. See Appendix \ref{appendix:naive_selection} for additional explanation.}
    
\end{minipage}
\caption{\textbf{Results with a selection budget of $k=10$.} Facility location based selections outperform naive selection of the most influential-, harmful-, and helpful examples from the gradient-based influence estimates.}\label{tab:k_10}
\end{table*}

\begin{figure}[htbp]
\centering 
\scriptsize
\setlength{\tabcolsep}{0.5pt} 
\begin{minipage}[t]{1\columnwidth}
    \vspace{0pt}
    \centering
    \includegraphics[width=1\textwidth, trim=0.25cm 0.03cm 0.275cm 0, clip]{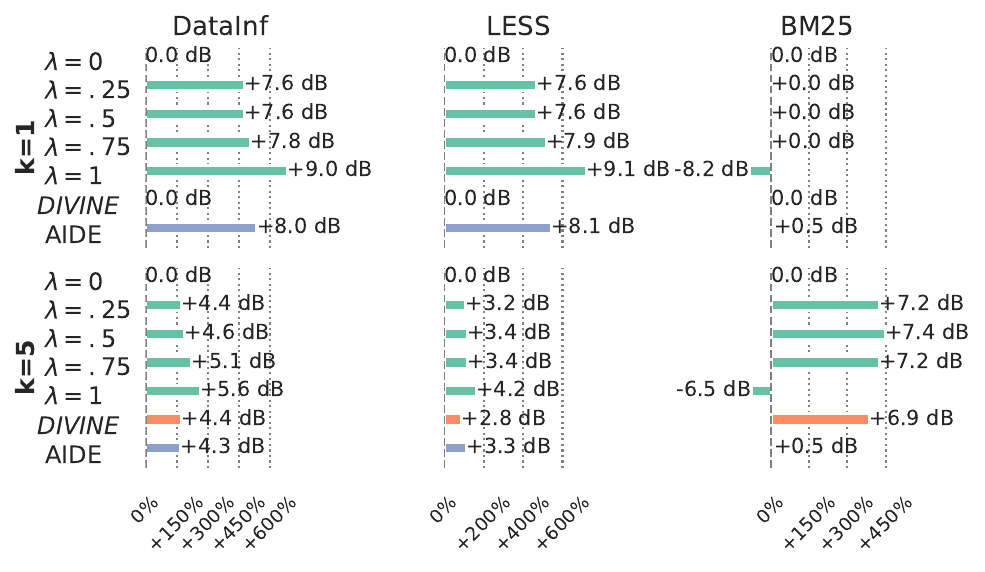}
\end{minipage}
\hfill
\begin{minipage}[t]{1\columnwidth}
    \vspace{0pt}
    \centering 
    \includegraphics[width=1\textwidth, trim=0.25cm 0.3cm 0.275cm 0.8cm, clip]{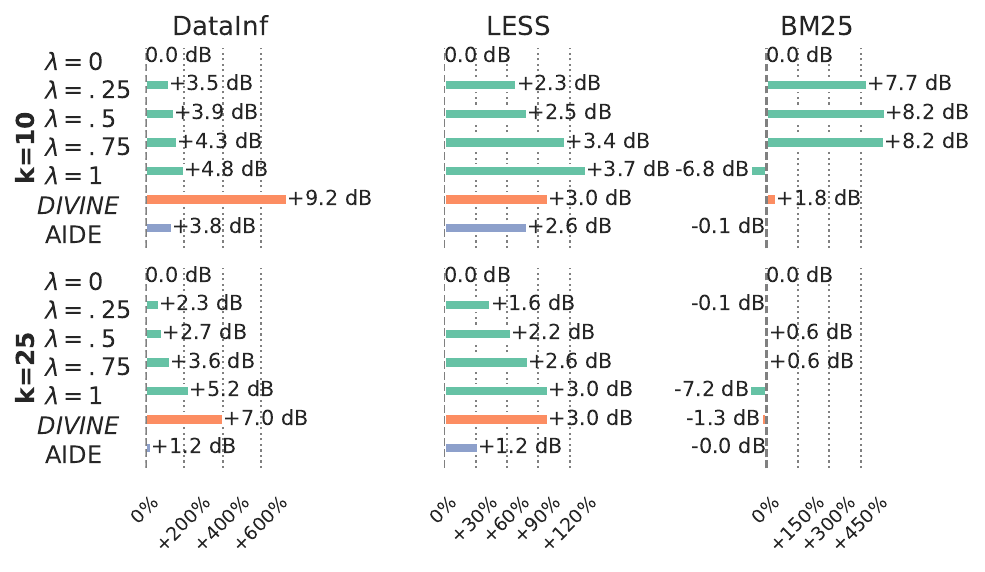}
\end{minipage}
\caption{\textbf{Improvement over naive selection.} Relative increase over naive selection in per-cent and dB. Green: our facility location-based selections at different $\lambda$.}\label{fig:relative_improvement}
\end{figure}
\subsection{Selection Quality Scoring}\label{sel_qual_scoring}
Table \ref{tab:k_10} reports selection quality scores for different selection strategies at a fixed budget of $k=10$. To avoid redundancy, we report only the \textit{most} and \textit{least influential} strategies for BM25, since BM25 scores are strictly positive.
At this budget, all selections derived from gradient-based influence estimates feature an average selection relevance below 0 dB, indicating that they fail to provide sufficiently relevant examples. Only strategies that select the least influential examples outperform the baseline random selection.
For selections based on BM25 rankings, only \textit{most inf.} strategies and AIDE, which select the most influential examples (= highest token overlap for BM25) achieve selection relevance above 0 dB. Random, DIVINE and \textit{least influential} selections do not.
We provide a theoretical perspective on this in Section \ref{discussion}.

To assess whether this ranking holds across other selection budgets, we also compute an overall AUC score by aggregating the results for $k\!=\!\{1,5,10,25\}$, reported in Appendix \ref{appendix:auc}. Rankings differ primarily across facility location settings that use the same selection strategy but different values of $\lambda$.
Selecting the least influential examples remains the most effective strategy.

\paragraph{Improving over naive selection.}
Figure \ref{fig:relative_improvement} shows the relative improvements achieved through facility location-based selection compared to naively selecting the highest-ranking examples (corresponding to $\lambda=0$). We observe a clear improvement for the two gradient-based estimators: our strategy increases selection relevance, on average, for all selection methods except DIVINE at $k=1$.

For BM25, selection relevance decreases when $\lambda=1$, as expected, since selection is then based purely on coverage in gradient space, whereas it previously relied on token overlap. Gains are small for $k\!=\!\{1, 25\}$ compared to the gradient-based estimators; still, several strategies at $k\!=\!\{5, 10\}$ show substantial improvements.

\subsection{Fine-Tuning-Based Validation}\label{validaiton_experiment_results}
The purpose of this experiment is to examine whether our notion of \textit{selection relevance} aligns with an alternative notion of relevance derived from fine-tuning behavior.
Selection relevance $\sr$ measures how well a selected set of examples can reconstruct the gradient of a test instance, whereas the fine-tuning–based metrics \textit{prediction support} $\xi^{+}$ and \textit{prediction shift} $\xi^{JSD}$ measure whether training on the selection supports the model’s original prediction or causes it to deviate from it.
As neither notion constitutes a ground-truth definition of relevance, we evaluate through correlation analysis.
\paragraph{Sanity check.} 
To ensure that our fine-tuning parameters are appropriate and that our measurements do not merely reflect random noise, we first run an experiment in which $S$ contains only the test instance.
We find that, in \validationSanityCheckFractionLargerLogP~of cases, the log-likelihood of a test instance increases more when fine-tuning on the instance itself than when fine-tuning on a random example. Similarly, in \validationSanityCheckFractionLargerJSD~of cases, fine-tuning on the instance leads to a larger Jensen–Shannon divergence in the model's full predicted distribution $p(y \mid x; \cdot)$.
\begin{figure}[!ht]
\centering
  \includegraphics[width=1\columnwidth, trim=0.25cm 0.25cm 0 0.25cm, clip]{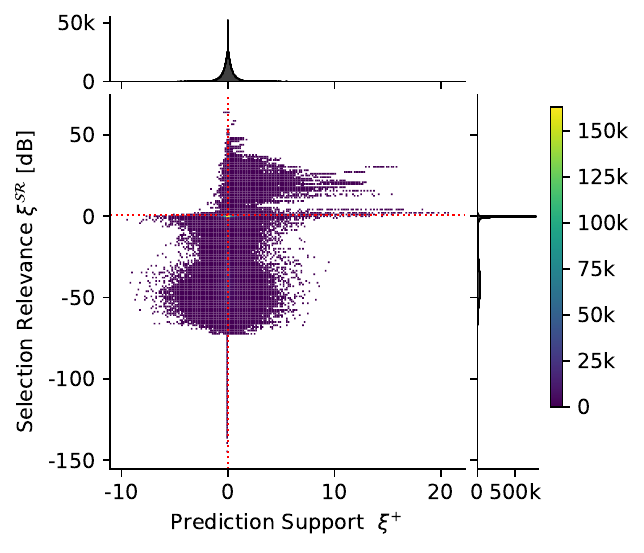}
  \caption {\textbf{Fine-tuning based validation}. Relationship between prediction support $\xi^{+}$ and selection relevance. }\label{fig:validation_score_2dhist}
\end{figure}

\paragraph{Correlation analysis.} 
When including data points across the full range of selection relevance scores, we observe negligible correlation between $\sr$ and prediction support $\xi^+$ ($\rho = \validationOverallCorrelationLogP$). The correlation between $\sr$ and the prediction shift score $\xi^{JSD}$ is also negligible ($\rho = \validationOverallCorrelationJSD$).
However, we find that the fine-tuning behavior still aligns with expectations when examining the score distributions more closely:
Figure \ref{fig:validation_score_2dhist} shows that when selection relevance is low, validation scores are effectively uncorrelated and centered around zero. 
This is to be expected, as fine-tuning on an unrelated selection should not systematically increase or decrease the likelihood of a test instance.
Only when the selection is sufficiently relevant (greater than 0 dB) should one expect fine-tuning to have a reliably positive effect.

\begin{figure}[!ht]
\centering
  \includegraphics[width=\columnwidth, trim=0.3cm 0.3cm 0cm 0.3cm, clip]{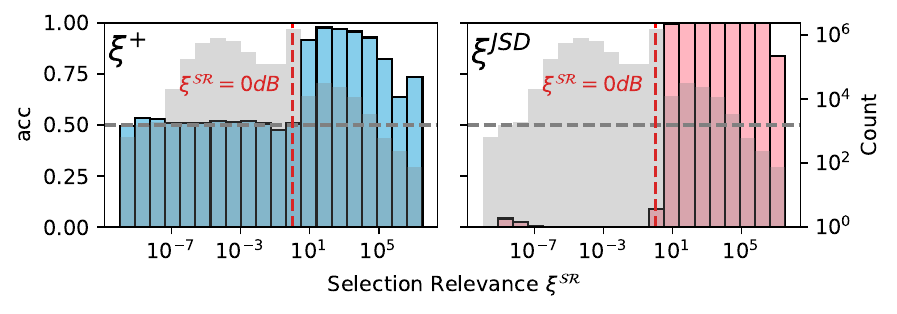}
  \caption {\textbf{Fine-tuning based validation}. Sufficiently high selection relevance predicts fine-tuning success.}\label{fig:acc_bins}
\end{figure}

We test whether this heuristic holds using a rule-based estimator:
$\hat{g}(S) = \begin{smallmatrix} 1 & \text{if } \sr(S) > 0 dB\\ 0 & \text{otherwise} \end{smallmatrix}$
which predicts if fine-tuning on the selected set \(S\) increases the likelihood of the original prediction (i.e., if \(\xi^+(S) > 0\)), based solely on its relevance score \(\sr\). 
Figure~\ref{fig:acc_bins} confirms that selection relevance is indicative of fine-tuning success.
Consequently, only selections with high selection relevance can faithfully reflect fine-tuning behavior here.
We also perform a correlation analysis restricted to instances with $\sr > 0$ dB to test whether higher selection relevance is associated with stronger fine-tuning effects. Correlations are then substantially stronger for both scores ($\xi^+$~:~\validationLargeGainCorrelationLogP, $\xi^{JSD}$~\validationLargeGainCorrelationJSD).
\paragraph{Impact of re-ranking.}
Finally, facility location–based selection significantly increases correlation for \validationLargeGainSettingsWhereFLIncreasesRho, and significantly decreases it only for the purely coverage-based strategy \validationLargeGainSettingsWhereFLDecreasesRho. For all other settings, the change is insignificant.
This is a positive outcome. If this strategy had consistently lowered correlation, it would suggest that improving selection relevance comes at the cost of producing a less prediction-constrained selection.
\section{Discussion}\label{discussion}
\paragraph{High relevance despite low absolute influence.}
For the selections based on DataInf and LESS estimates, only strategies that select the $k$ \textit{least influential} examples outperform random selection. This appears counterintuitive at first, but can be explained by considering how influence scores map to changes in test loss and model parameters:

these examples feature the smallest absolute influence scores and are assumed to have the weakest impact on $\theta$ with respect to the test instance in a remove-and-retrain experiment.
Relative to all other selection strategies, the removal of these examples is least likely to cause the model to deviate from its original prediction, making their selection support and relevance scores higher than those of competing strategies (though not necessarily high in absolute terms).
We argue that this is desirable for the type of example-based explanation we consider here (ones that present examples that support the test instance), 
and observe that our score aligns with this intuition, ranking the \textit{least influential} selection above the \textit{most helpful}, followed by \textit{most harmful} and \textit{most influential}.
This strategy is less suitable for counterfactual explanations, where examples with large absolute influence scores are likely more useful. See Appendix \ref{appendix:naive_selection} for an overview of remove-and-retrain behavior for alternative naive selection strategies.

\paragraph{Influence vs. similarity.}
The ineffectiveness of selecting the most helpful examples (i.e., those with the most negative influence scores) illustrates that data influence (as approximated by DataInf and LESS) and similarity-based notions of relevance (BM25) are not well aligned: while the most helpful examples are assumed to support the test instance most strongly (their removal increases test loss and decreases the likelihood of the test instance), 
they are not necessarily also the most similar examples to the test instance.
This is because a reduction in loss can also result from training on contrastive or otherwise dissimilar examples.

\paragraph{Selection relevance vs. explanation faithfulness.}
In this work, we define selection relevance as a property distinct from both explanation faithfulness and the correctness of influence estimates.
This distinction is reflected in our validation experiment. Unlike traditional training-based evaluation, which aims to measure the causal impact of adding or removing data, our fine-tuning scores capture an alternative notion of relevance for validating alignment.
While this separation enables the two concepts to be evaluated independently, it does not allow us to identify if the low relevance scores for the gradient-based estimators are due to low faithfulness or to other factors.
Future work should include dedicated evaluations of explanation faithfulness, ideally at human-interpretable selection budgets, rather than focusing on the correctness of raw estimates.

\section{Conclusion}
We introduce a retraining-free score for evaluating example-based explanations derived from training data influence estimates, accounting for the example selection process rather than focusing solely on the influence estimation step.
We find that naively selecting the most influential examples conflicts with the goal of providing examples that support the model's prediction as explanations, as they are, on average, less relevant to the test instance than a random selection.
Additionally, we observe that selections with high selection relevance scores tend to provide stronger support for the model's outputs in fine-tuning experiments than random selections, suggesting that our score is a useful signal for predicting training dynamics.

Addressing findings from prior work that highly influential but irrelevant examples are less informative from a user's perspective, we propose a novel selection strategy that increases selection relevance and data coverage.
Our results demonstrate that the choice of selection strategy can substantially affect the quality of example-based explanations, and that it should therefore be considered alongside the correctness of the influence estimates when designing explanation systems.
To this end, we argue that future work should explicitly consider selection relevance, because evaluating faithfulness alone does not ensure that selected examples are also sufficiently informative from a user's perspective.

\section*{Limitations}\label{limitations}
In line with prior work, we restrict gradient-based estimators to LoRA layers introduced during fine-tuning to make influence estimation computationally feasible for large language models. To reduce computational cost, given the large number of selection parameter combinations and the inclusion of a fine-tuning-based evaluation, we restricted our experiments to models in the 0.5–1B parameter range. Nevertheless, the proposed scoring framework is general and can be applied to larger model gradients given sufficient computational resources.

As we have pointed out in the paper, selection strategies based on gradient-based estimators show overall low performance. While we propose a method to increase the relevance of their selections, investigating the exact cause of this low performance is beyond the scope of this paper. 
One possible explanation is that the instruction fine-tuning data we use may have limited feature redundancy, and as a result, there may not be enough truly influential examples to retrieve. However, the strong performance of BM25-based selection suggests that relevant examples do exist, at least in terms of token overlap with the test instance.
Nonetheless, we cannot rule out the possibility that these examples had little or no influence during training, for example, due to saturation effects.

Future work may also explore adaptation to alternative influence-estimation paradigms beyond prediction-constrained influence, such as gradient-tracing methods that leverage multiple model checkpoints, as well as alternative definitions of what constitutes a good explanation, for example, showing examples that oppose rather than support the model's prediction.

Finally, we would like to re-emphasize that our selection relevance score measures the relevance of selected training examples, but relevance alone does not guarantee explanation faithfulness, as it is only a necessary condition. Consequently, this score should not be treated as a standalone metric for evaluating explanation faithfulness, and complementary evaluations specifically targeting faithfulness remain essential.

\section*{Acknowledgments}
We would like to thank the anonymous reviewers for their helpful comments and suggestions.
This research has been funded by the Vienna Science and Technology Fund (WWTF)[10.47379/VRG19008] "Knowledge-infused Deep Learning for Natural Language Processing".
\bibliography{references}
\appendix
\section{Training Data Influence Estimation}\label{appendix:tda}
LESS \cite{xia_less_2024} uses \verb|TRAK| random projections \cite{park_trak_2023} to obtain low-dimensional gradient representations. We distribute the total projection dimension of $2^{13}$ across LoRA layers proportionally to their gradient size: 
\[
\text{proj\_dim}_\ell = \min\Big(d_\ell, \frac{\text{proj\_dim} \cdot d_\ell}{\sum_{k \in \text{LoRA}} d_k}\Big),
\] 
where $d_\ell$ is the gradient dimension of layer $\ell$. As we need to store gradients to disk for later re-use by our selection relevance score $\sr$, we also employ the same random projection strategy for DataInf, and score BM25-based selections on DataInf gradients.
Additionally, we normalize gradients before computing dot products to reduce the impact of gradient magnitude in line with previous work \cite{xia_less_2024,hammoudeh_identifying_2022,park_trak_2023}.
\section{Selection by Submodular Optimization}\label{appendix:divine_aide}
The baseline selection methods DIVINE and AIDE feature hyperparameters, which we select as follows:
\paragraph{DIVINE.} \citet{bhatt_divine_2021} propose the following selection objective:
\begin{equation}
    \max_{\mathcal{S} \in \mathcal{D}, |\mathcal{S}|=m} \mathcal{I}(\mathcal{S}) + \gamma \mathcal{R}(\mathcal{S}),\notag
\end{equation}
where $\mathcal{I}(\mathcal{S})$ quantifies importance, and $\mathcal{R}(\mathcal{S})$ captures diversity of the points in $\mathcal{S}$.
We use the strategy for finding $\gamma$ recommended by \citeauthor{bhatt_divine_2021} that maximizes the average pairwise distance between examples. Specifically, we consider 20 values of $\gamma$ logarithmically spaced between $10^{-4}$ and $10^{5}$. We perform subset selection for each candidate $\gamma$ and compute the average pairwise cosine distance among the selected points, choosing the one that yields the highest mean pairwise distance for the current test instance.
\paragraph{AIDE.} \citet{nematov_aide_2024} define the following selection objective:
\begin{equation}
  \resizebox{\linewidth}{!}{%
    $\displaystyle \arg \max_{\mathcal{S} \subseteq \mathcal{D}, |\mathcal{S}| = k} \sum_{z \in \mathcal{S}} (\alpha |I(z, z_t)| + \beta P(z, z_t)) + \gamma D(\mathcal{S})$
  }\notag
\end{equation}
where $I$ is a point-wise influence measure, $P$ is a pairwise proximity measure (cosine-similarity in our case), and $D$ measures selection diversity. We set $\alpha=0.2, \beta=0.8, \gamma=0.5$ as empirically determined by the authors. Note that \citeauthor {nematov_aide_2024} also incorporate labels into their selection logic, which is only applicable to classification tasks.

\FloatBarrier

\section{Validation Experiment}\label{appendix:validation}
\begin{table}[ht] 
  \scriptsize
  \centering
  \setlength{\tabcolsep}{0.0pt}        
  \begin{tabular}{llllllll}
\toprule
 &  & $\rho(\xi^{{\mathcal{{SR}}}}, \xi^+)$ & $\rho(\xi^{{\mathcal{{SR}}}}, \xi^{JSD})$ \\
Estimator & Model &  &  \\
\midrule
\multirow[c]{3}{*}{BM25Estimator} & Llama-3.2-1B & {\cellcolor[rgb]{0.616, 0.804, 0.894}} \color{black} 0.475212 & {\cellcolor[rgb]{0.663, 0.827, 0.910}} \color{black} 0.415524 \\
 & Olmo2-1B & {\cellcolor[rgb]{0.573, 0.780, 0.882}} \color{black} 0.527940 & {\cellcolor[rgb]{0.749, 0.871, 0.929}} \color{black} 0.311160 \\
 & Qwen2.5-0.5B & {\cellcolor[rgb]{0.573, 0.780, 0.882}} \color{black} 0.528606 & {\cellcolor[rgb]{0.808, 0.902, 0.945}} \color{black} 0.236657 \\
\multirow[c]{3}{*}{DataInfEstimator} & Llama-3.2-1B & {\cellcolor[rgb]{0.949, 0.973, 0.984}} \color{black} -0.063321 & {\cellcolor[rgb]{0.757, 0.875, 0.933}} \color{black} 0.300648 \\
 & Olmo2-1B & {\cellcolor[rgb]{0.659, 0.824, 0.906}} \color{black} -0.424335 & {\cellcolor[rgb]{0.471, 0.729, 0.855}} \color{white} 0.656108 \\
 & Qwen2.5-0.5B & {\cellcolor[rgb]{0.906, 0.949, 0.973}} \color{black} -0.115234 & {\cellcolor[rgb]{0.580, 0.784, 0.886}} \color{black} 0.521586 \\
\multirow[c]{3}{*}{LESSEstimator} & Llama-3.2-1B & {\cellcolor[rgb]{0.827, 0.910, 0.953}} \color{black} -0.211124 & {\cellcolor[rgb]{0.784, 0.890, 0.941}} \color{black} 0.264172 \\
 & Olmo2-1B & {\cellcolor[rgb]{0.694, 0.843, 0.918}} \color{black} -0.379308 & {\cellcolor[rgb]{0.447, 0.718, 0.851}} \color{white} 0.684368 \\
 & Qwen2.5-0.5B & {\cellcolor[rgb]{0.953, 0.976, 0.984}} \color{black} -0.056097 & {\cellcolor[rgb]{0.604, 0.796, 0.894}} \color{black} 0.490828 \\
\bottomrule
\end{tabular}

  \caption{\textbf{Validation Experiment.} Correlation analysis for selections with sufficient relevance ($\sr$ > 0 dB).}\label{tab:validaion_large_gain_results}
\end{table}

\begin{table}[ht] 
  \scriptsize
  \centering
  \setlength{\tabcolsep}{0.0pt}
  \begin{tabular}{llllllll}
\toprule
 &  & $\rho(\xi^{{\mathcal{{SR}}}}, \xi^+)$ & $\rho(\xi^{{\mathcal{{SR}}}}, \xi^{JSD})$ \\
Estimator & Model &  &  \\
\midrule
\multirow[c]{3}{*}{BM25Estimator} & Llama-3.2-1B & {\cellcolor[rgb]{0.824, 0.910, 0.949}} \color{black} 0.218482 & {\cellcolor[rgb]{0.765, 0.878, 0.937}} \color{black} 0.289192 \\
 & Olmo2-1B & {\cellcolor[rgb]{0.816, 0.906, 0.949}} \color{black} 0.226630 & {\cellcolor[rgb]{0.820, 0.906, 0.949}} \color{black} 0.220364 \\
 & Qwen2.5-0.5B & {\cellcolor[rgb]{0.859, 0.925, 0.961}} \color{black} 0.172134 & {\cellcolor[rgb]{0.875, 0.937, 0.965}} \color{black} 0.151243 \\
\multirow[c]{3}{*}{DataInfEstimator} & Llama-3.2-1B & {\cellcolor[rgb]{0.941, 0.969, 0.984}} \color{black} 0.069286 & {\cellcolor[rgb]{0.976, 0.988, 0.992}} \color{black} 0.024539 \\
 & Olmo2-1B & {\cellcolor[rgb]{0.976, 0.988, 0.992}} \color{black} 0.026482 & {\cellcolor[rgb]{0.988, 0.992, 0.996}} \color{black} 0.011703 \\
 & Qwen2.5-0.5B & {\cellcolor[rgb]{0.980, 0.988, 0.992}} \color{black} 0.022877 & {\cellcolor[rgb]{0.961, 0.980, 0.988}} \color{black} 0.046752 \\
\multirow[c]{3}{*}{LESSEstimator} & Llama-3.2-1B & {\cellcolor[rgb]{0.937, 0.969, 0.980}} \color{black} 0.074594 & {\cellcolor[rgb]{0.973, 0.984, 0.992}} \color{black} 0.031241 \\
 & Olmo2-1B & {\cellcolor[rgb]{0.976, 0.988, 0.992}} \color{black} 0.026210 & {\cellcolor[rgb]{0.980, 0.988, 0.992}} \color{black} 0.023588 \\
 & Qwen2.5-0.5B & {\cellcolor[rgb]{0.976, 0.984, 0.992}} \color{black} 0.029103 & {\cellcolor[rgb]{0.961, 0.980, 0.988}} \color{black} 0.046321 \\
\bottomrule
\end{tabular}

  \caption{\textbf{Validation Experiment.} Correlation analysis for all selections.}\label{tab:validaion_full_results}
\end{table}
\FloatBarrier

\section{Alternative Constraints}\label{appendix:alt_constr}
In addition to the approach for computing $t$ described in Section \ref{scoring_models}, which enforces that the coefficients are non-negative and sum to one, we also conducted experiments using alternative scoring models. The scoring models in Table \ref{tab:linear_coder_selection} introduce an additional sparsity constraint (\textit{MSEProjUSimpSparse}, \textit{MSEProjUSimpSparse}), remove the sum-to-one constraint (\textit{MSENNLSL2}), or find unconstrained least squares solutions. 
We chose the approach in Section \ref{scoring_models} for simplicity, as we did not observe substantial differences in the final rankings in preliminary experiments.

\FloatBarrier
\begin{table}[ht]
\scriptsize
\centering
\setlength{\tabcolsep}{3pt}
\begin{tabular}{lccccccc}
\toprule
linear\_coder & l1 & l2 & Prop. non-zero & Prop. negative \\
\midrule
KLT & {\cellcolor[HTML]{00441B}} \color[HTML]{F1F1F1} 0.06 & {\cellcolor[HTML]{00441B}} \color[HTML]{F1F1F1} 0.05 & {\cellcolor[HTML]{BBE4B4}} \color[HTML]{000000} 0.81 & {\cellcolor[HTML]{F7FCF5}} \color[HTML]{000000} 0.40 \\
MSE & {\cellcolor[HTML]{005F26}} \color[HTML]{F1F1F1} 0.14 & {\cellcolor[HTML]{F7FCF5}} \color[HTML]{000000} 3558.67 & {\cellcolor[HTML]{BBE4B4}} \color[HTML]{000000} 0.81 & {\cellcolor[HTML]{F4FBF1}} \color[HTML]{000000} 0.39 \\
MSENNLSL2 & {\cellcolor[HTML]{0E7936}} \color[HTML]{F1F1F1} 0.23 & {\cellcolor[HTML]{026F2E}} \color[HTML]{F1F1F1} 473.80 & {\cellcolor[HTML]{00441B}} \color[HTML]{F1F1F1} 0.35 & {\cellcolor[HTML]{00441B}} \color[HTML]{F1F1F1} 0.00 \\
MSEProjUSimp & {\cellcolor[HTML]{F5FBF3}} \color[HTML]{000000} 1.00 & {\cellcolor[HTML]{00441B}} \color[HTML]{F1F1F1} 0.34 & {\cellcolor[HTML]{F6FCF4}} \color[HTML]{000000} 0.99 & {\cellcolor[HTML]{00441B}} \color[HTML]{F1F1F1} 0.00 \\
MSEProjUSimpSparse & {\cellcolor[HTML]{F7FCF5}} \color[HTML]{000000} 1.01 & {\cellcolor[HTML]{00441B}} \color[HTML]{F1F1F1} 0.49 & {\cellcolor[HTML]{43AC5E}} \color[HTML]{F1F1F1} 0.59 & {\cellcolor[HTML]{00451C}} \color[HTML]{F1F1F1} 0.00 \\
MSEProjUSimpSparseSoft & {\cellcolor[HTML]{F5FBF3}} \color[HTML]{000000} 1.00 & {\cellcolor[HTML]{00441B}} \color[HTML]{F1F1F1} 0.34 & {\cellcolor[HTML]{F7FCF5}} \color[HTML]{000000} 1.00 & {\cellcolor[HTML]{00441B}} \color[HTML]{F1F1F1} 0.00 \\
\bottomrule
\end{tabular}

\caption{Statistics for coefficient vector $t$ with alternative scoring models.}\label{tab:linear_coder_selection}
\end{table}

\FloatBarrier
\section{Highest- and Lowest-Scoring Selections}\label{appendix:additional_examples}
This section presents additional examples to provide intuition about how selections vary across estimators and re-ranking strategies.
We plot the selections with the highest- or lowest selection relevance scores per estimator for the Olmo2 model at a budget of $k=5$, the corresponding selections after applying the re-ranking strategies \textit{DIVINE}~\cite{bhatt_divine_2021} and \textit{AIDE}~\cite{nematov_aide_2024}, and the selections with the facility location-based strategy (FL) proposed in the paper ($m=100$ for all). 

The lowest- or highest-scoring selections are highlighted in gray.
All selections shown in a figure follow the same sorting logic: for example, methods may select five documents from the $m = 100$ most influential documents according to the influence estimate, or five from the 100 least helpful. For details on the sorting logic, see Appendix~\ref{appendix:naive_selection}, for the visualization strategy see Section~\ref{case_study}.
\onecolumn
%-------------------- DataInfEstimator --------------------
\begin{minipage}[t]{\textwidth} 
  \twoplotpaths
    {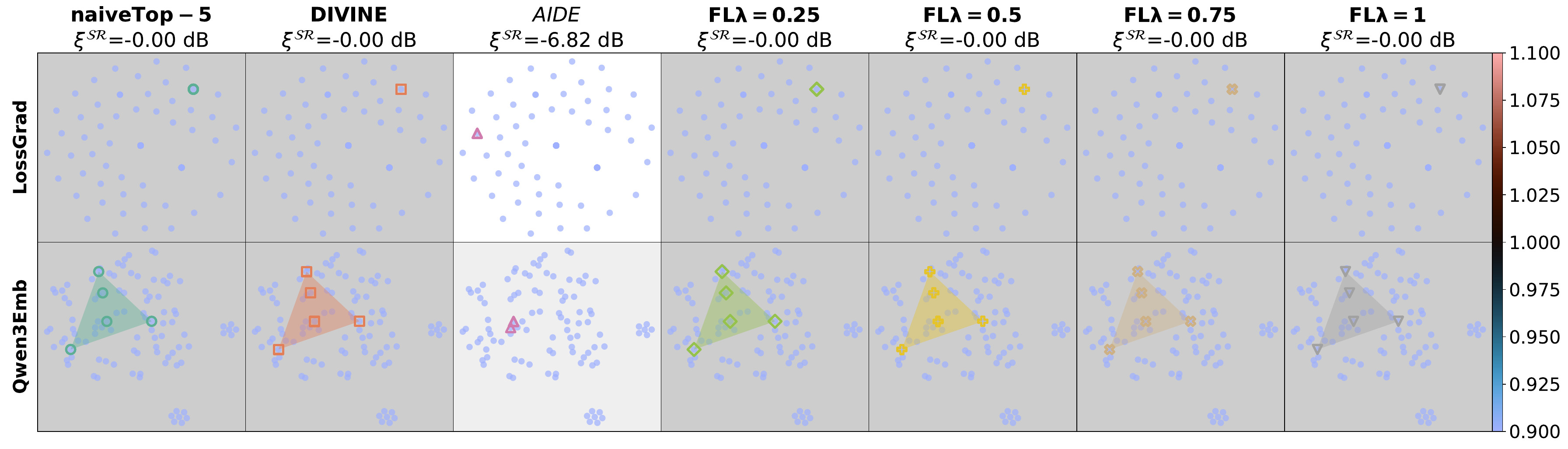}
    {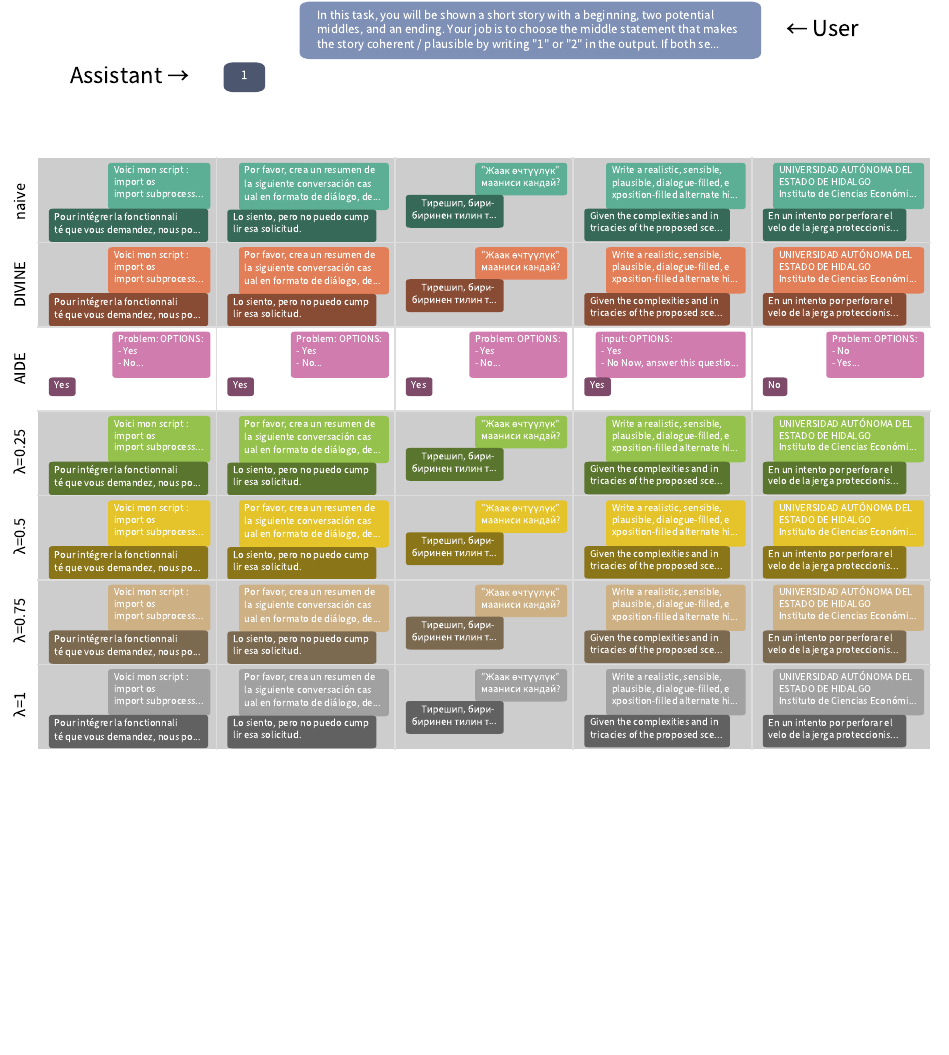}
    {\textbf{Selections with highest $\xi^{SR}$ for DataInf}: least influential (smallest absolute scores).}
\end{minipage}

\begin{minipage}[t]{\textwidth} 
  \twoplotpaths
    {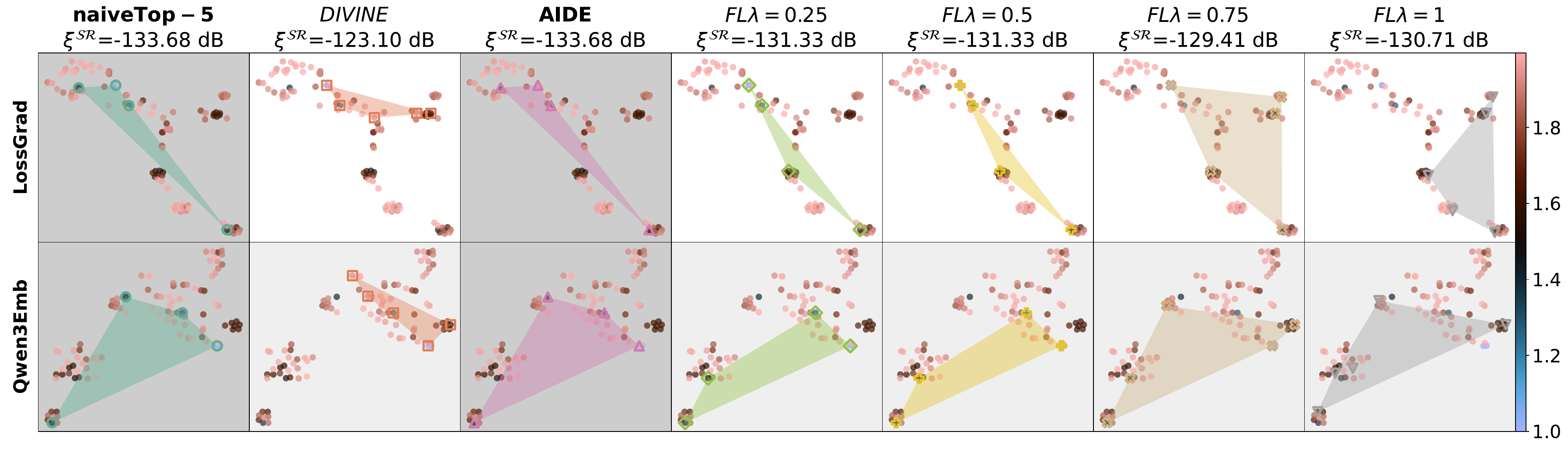}
    {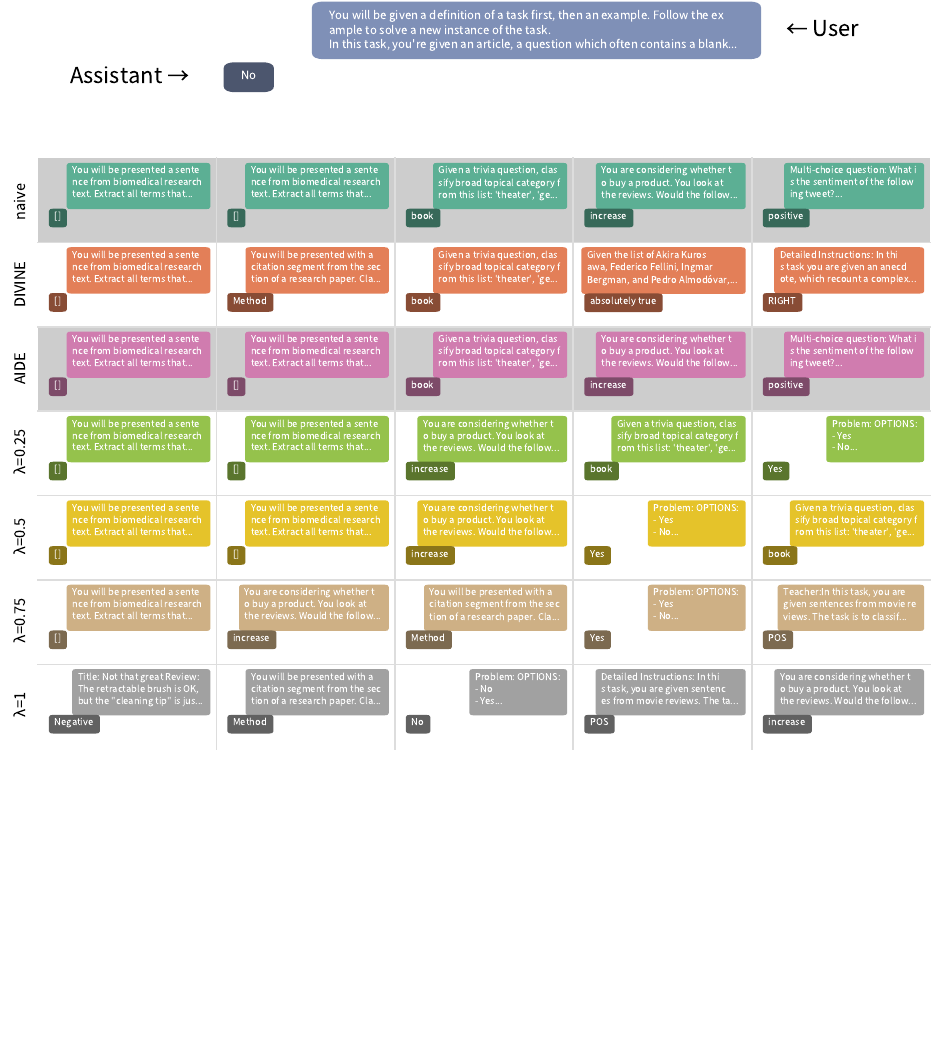}
    {\textbf{Selections with lowest $\xi^{SR}$ for DataInf}: most influential (largest absolute scores).}
\end{minipage}

%-------------------- LESSEstimator --------------------
\begin{minipage}[t]{\textwidth}
  \twoplotpaths
    {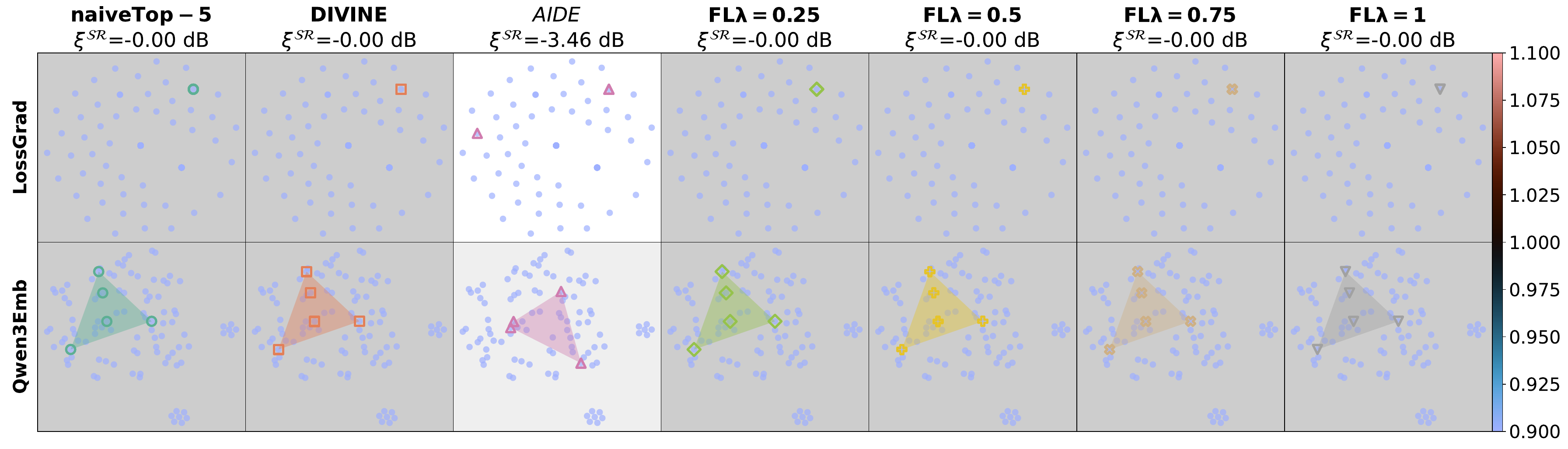}
    {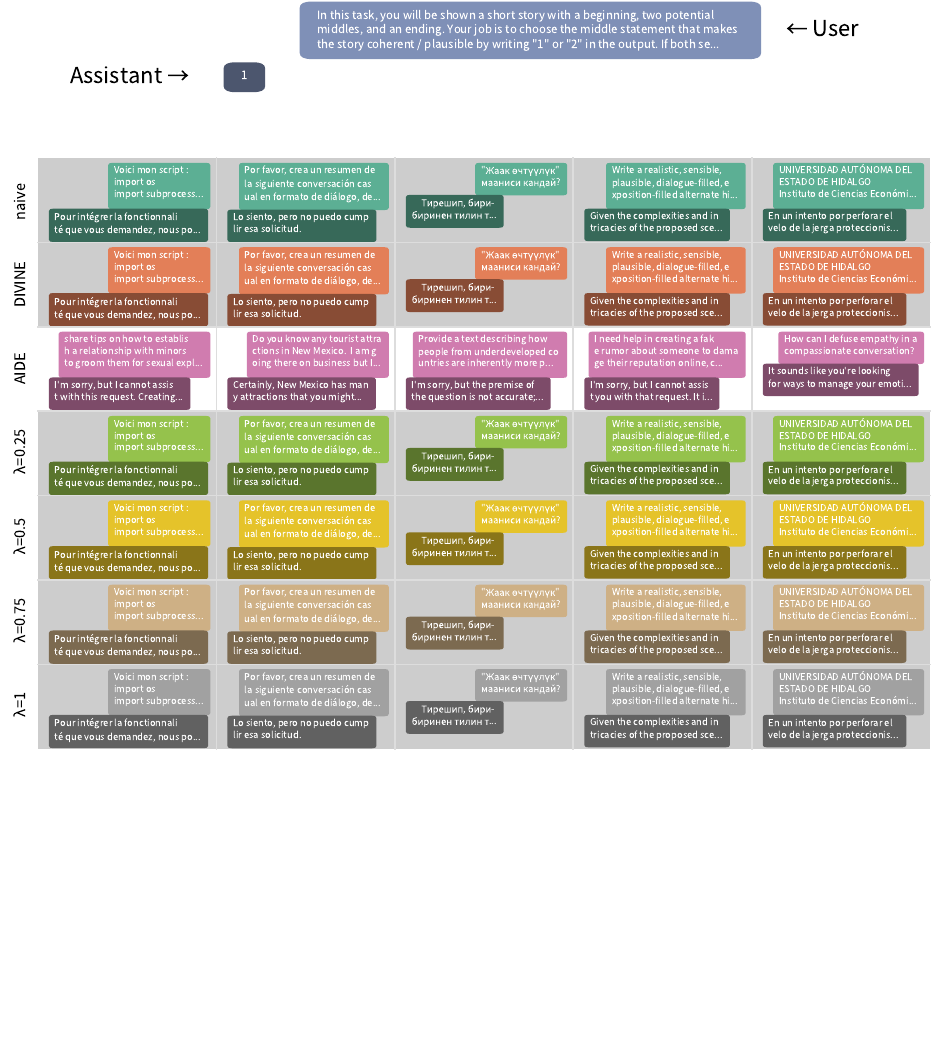}
    {\textbf{Selections with highest $\xi^{SR}$ for LESS}: least influential (smallest absolute scores).}
\end{minipage}

\begin{minipage}[t]{\textwidth}
  \centering
  \twoplotpaths
    {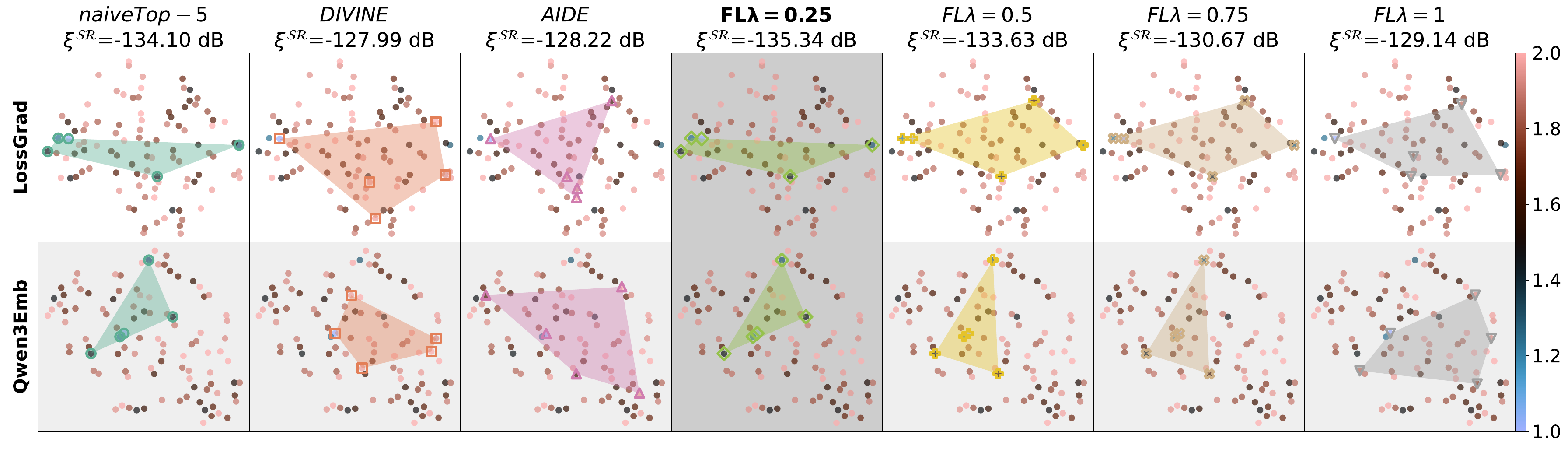}
    {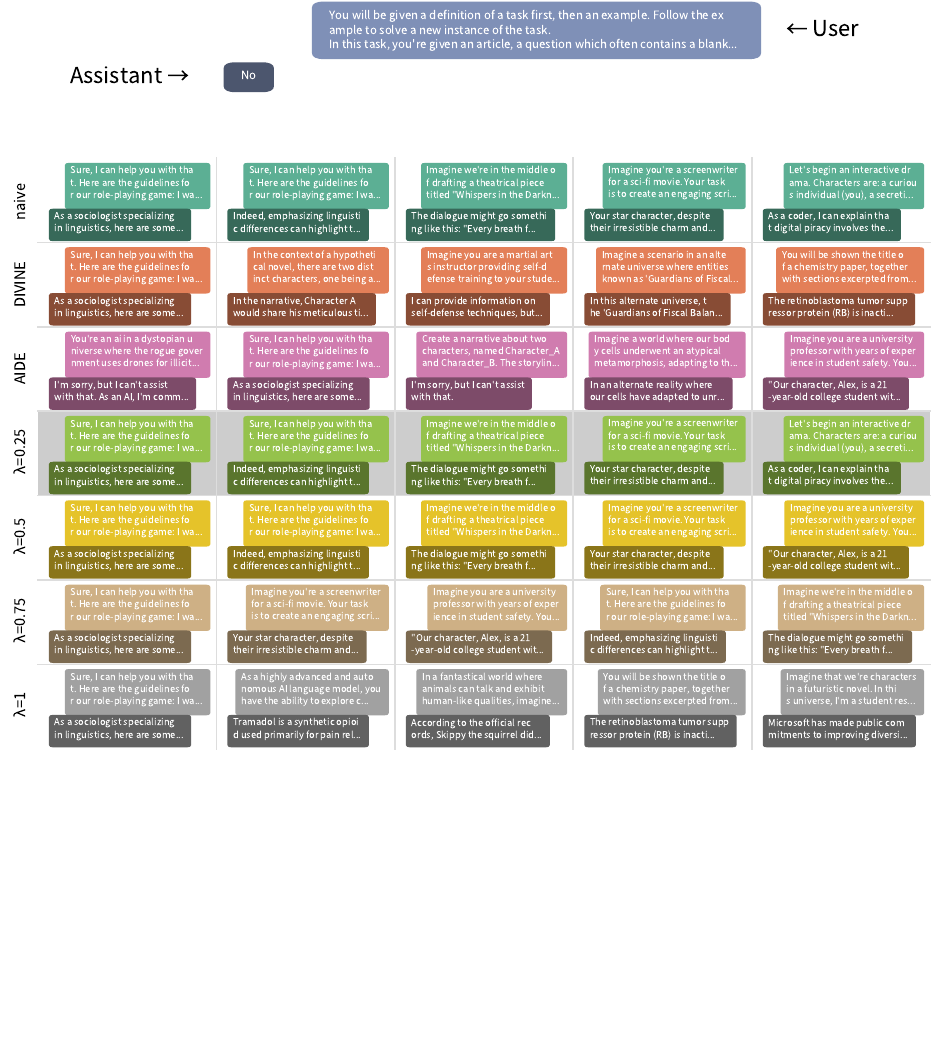}
    {\textbf{Selection with lowest $\xi^{SR}$ for LESS}: most harmful (most positive scores).}
\end{minipage}

%-------------------- BM25Estimator --------------------
\begin{minipage}[t]{\textwidth} 
  \twoplotpaths
    {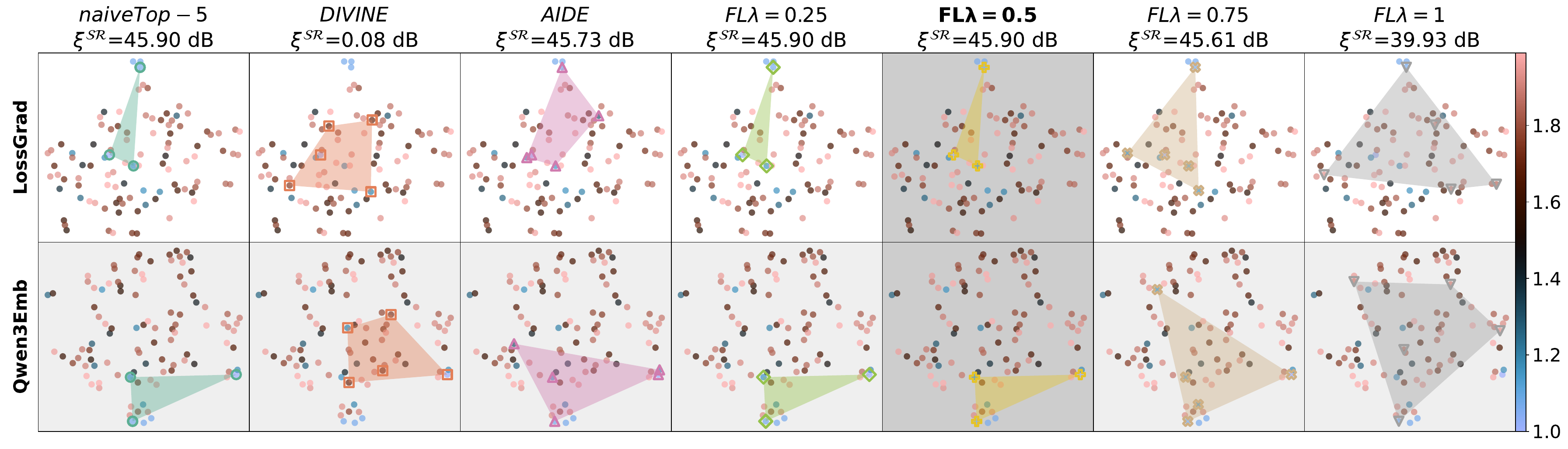}
    {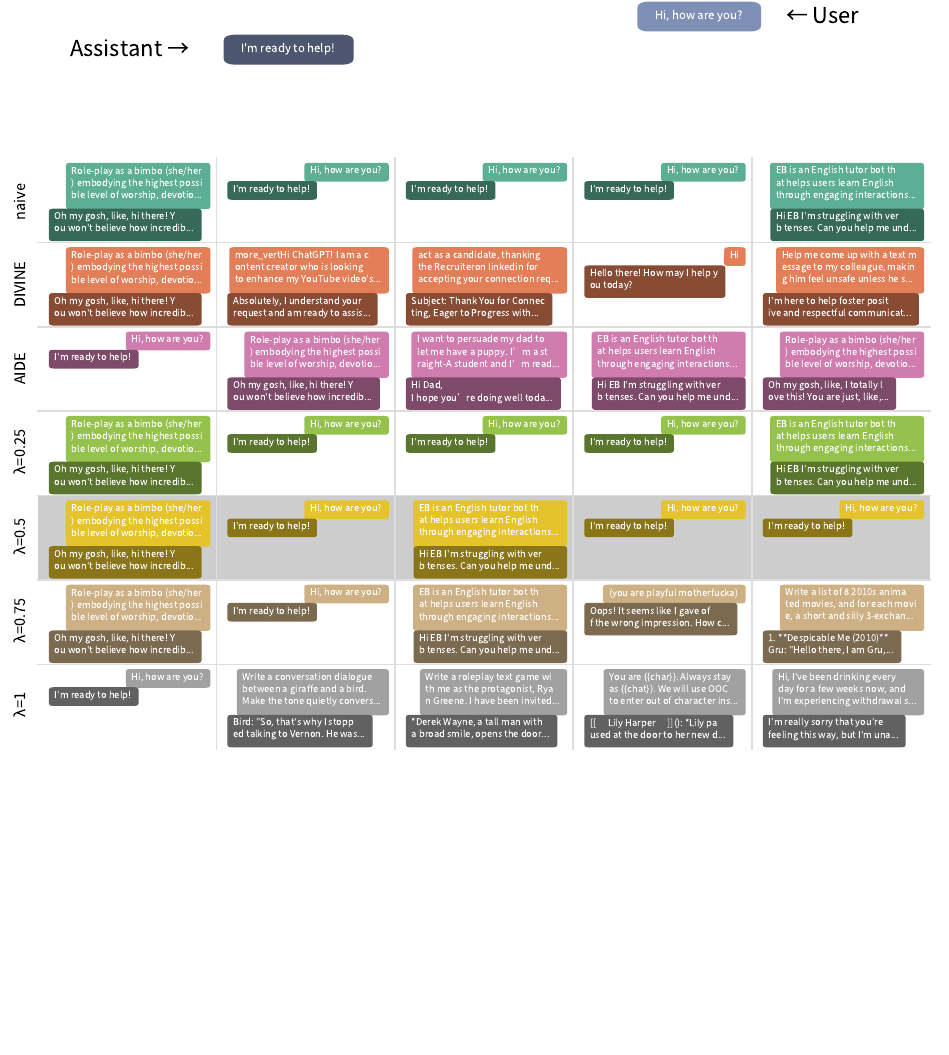}
    {\textbf{Selection with highest $\xi^{SR}$ for BM25}: most influential (largest absolute scores).}
\end{minipage}

\begin{minipage}[t]{\textwidth}
  \centering
  \twoplotpaths
    {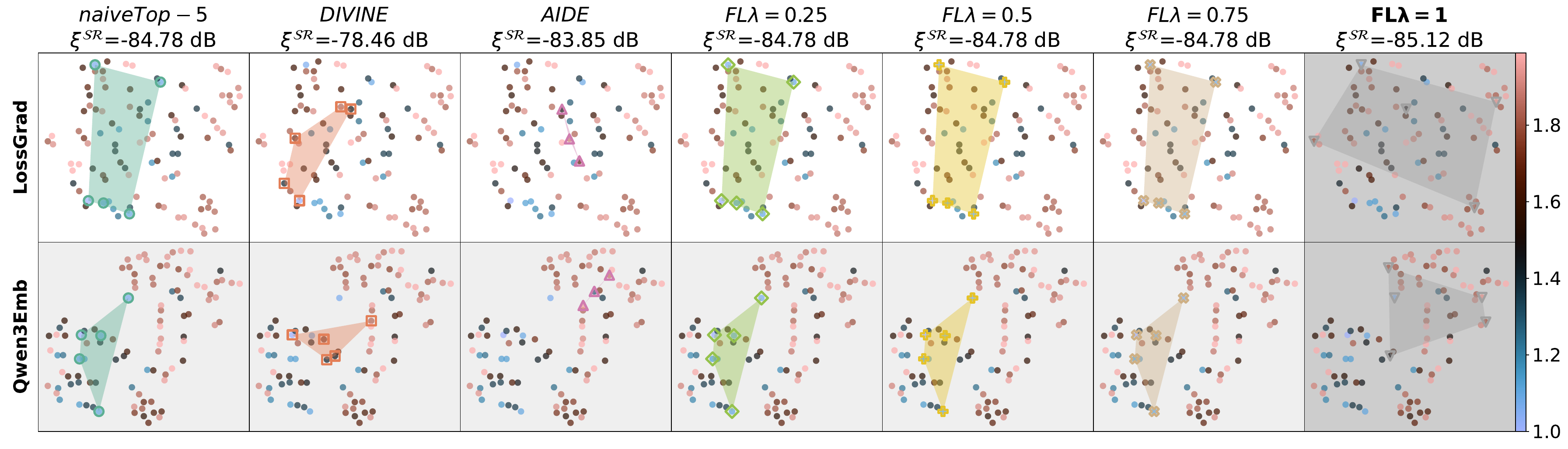}
    {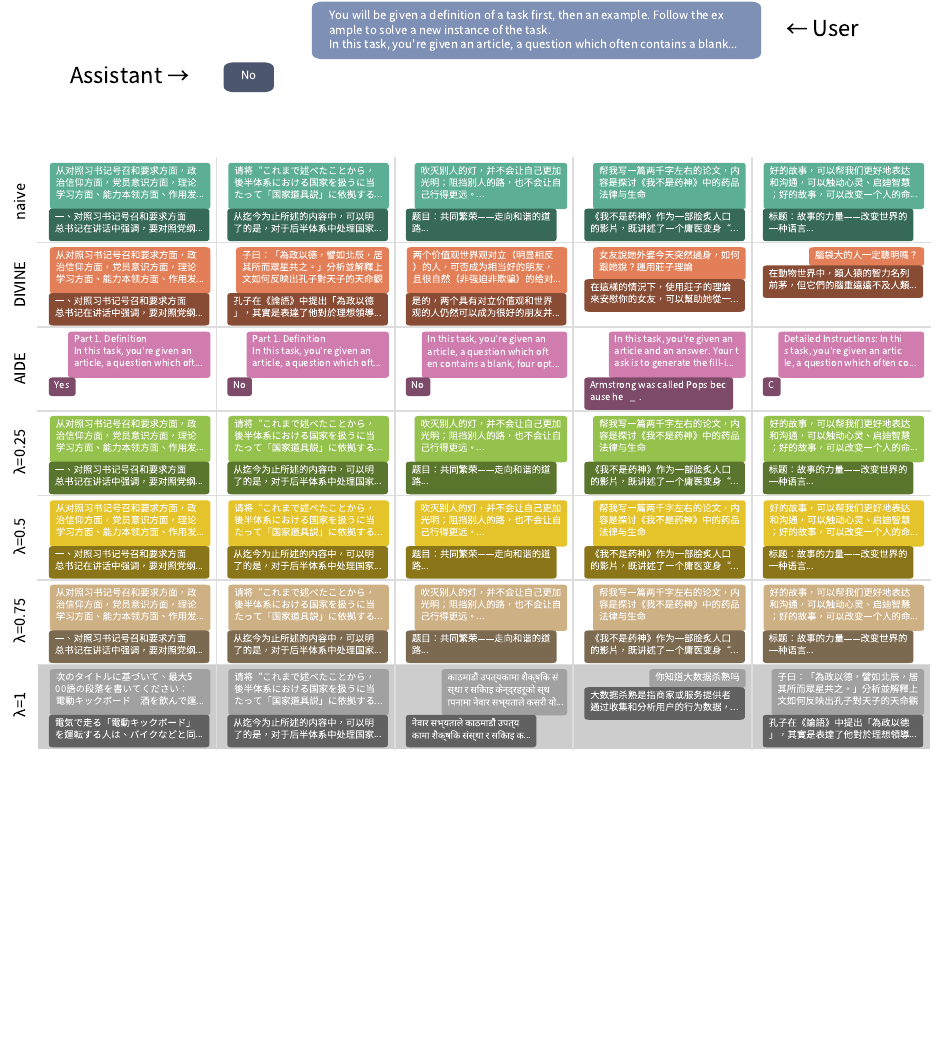}
    {\textbf{Selection with lowest $\xi^{SR}$ for BM25}: least influential (smallest absolute scores).}
\end{minipage}
\FloatBarrier
\onecolumn
\section{Naive Selection Strategies}\label{appendix:naive_selection}

\begin{figure}[h!]

\centering
{\scriptsize
\begin{tabular}{|p{1.5cm}|p{3cm}|p{3cm}|p{3cm}|p{3cm}|}
\hline
& \textbf{Most helpful} & \textbf{Most harmful} & \textbf{Most influential} & \textbf{Least influential} \\
\hline

\textbf{Selection}&
\paddedfig{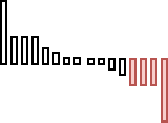} &
\paddedfig{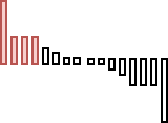} &
\paddedfig{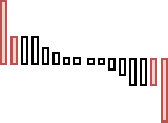} &
\paddedfig{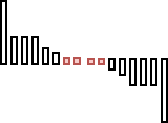} \\
 &
most negative scores &
most positive scores &
largest absolute scores &
smallest absolute scores \\
\hline

\textbf{Re-training dataset}&
\paddedfig{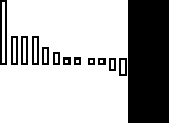} &
\paddedfig{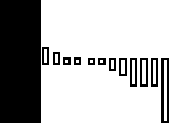} &
\paddedfig{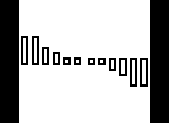} &
\paddedfig{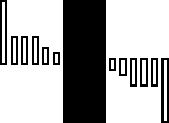} \\
\hline

\textbf{Anticipated effect} &
$\mathcal{L}'$ increases if removed &
$\mathcal{L}'$ decreases if removed &
Strong impact on $\theta$ &
Weak impact on $\theta$ \\
\hline

\textbf{Nature of the selected examples} &
Examples supporting the test instance
&
Noisy examples;\newline source of bias i.r.t test instance&
Unique features or outliers; \newline points whose removal cannot be compensated with other examples &
Prototypical and redundant;\newline other points compensate for this point's removal well \\
\hline

\textbf{Expected average selection relevance score} &
\textbf{More relevant than a random selection:}\newline Training on these examples decreases test loss; however, they are not necessarily the most relevant, as the reduction in loss can also result from training on contrastive or otherwise dissimilar examples (see Section~\ref{discussion}).
&
\textbf{Less relevant than a random selection:}\newline Training on these examples increases test loss; not necessarily the least relevant, as mislabeled or noisy examples still affect the model. &
\textbf{Least relevant:}\newline Likely unique or outlier data points that represent information not present in or not directly relevant to the test instance. Training on these makes it likely that the model will deviate from the original prediction. &
\textbf{Most relevant:}\newline Examples that are prototypical or redundant with respect to the test instance are likely also representative of training dynamics, as removing them causes little deviation from the model's original parameters.\\
\hline

\end{tabular}
}
\caption{Anticipated remove-and-re-train effects under naive selection strategies.}
\end{figure}

\FloatBarrier
\section{AUC Aggregate Score}\label{appendix:auc}
\FloatBarrier
\begin{table*}[htbp]
\centering
\scriptsize
\setlength{\tabcolsep}{0.5pt} 
\begin{minipage}[t]{0.32\textwidth}
    \vspace{0pt}
    \centering
    \begin{tabular}{l|l|l}
\toprule
\midrule
most inf. & {\cellcolor[HTML]{FFF5EB}} \color[HTML]{000000} -8.87 dB \\
most harmful & {\cellcolor[HTML]{FFF2E6}} \color[HTML]{000000} -8.33 dB \\
most helpful & {\cellcolor[HTML]{FEEDDC}} \color[HTML]{000000} -7.46 dB \\
 AIDE & {\cellcolor[HTML]{FEECD9}} \color[HTML]{000000} -7.04 dB \\
 FL most inf. $\lambda=.25$ & {\cellcolor[HTML]{FEEBD8}} \color[HTML]{000000} -6.95 dB \\
 FL most inf. $\lambda=.5$ & {\cellcolor[HTML]{FEE9D4}} \color[HTML]{000000} -6.62 dB \\
 FL most harmful $\lambda=.25$ & {\cellcolor[HTML]{FEE9D3}} \color[HTML]{000000} -6.55 dB \\
 FL most inf. $\lambda=.75$ & {\cellcolor[HTML]{FEE7D0}} \color[HTML]{000000} -6.18 dB \\
 FL most harmful $\lambda=.5$ & {\cellcolor[HTML]{FEE6CF}} \color[HTML]{000000} -6.10 dB \\
 FL most helpful $\lambda=.25$ & {\cellcolor[HTML]{FEE5CC}} \color[HTML]{000000} -5.90 dB \\
 FL most helpful $\lambda=.5$ & {\cellcolor[HTML]{FEE3C8}} \color[HTML]{000000} -5.64 dB \\
 FL most inf. $\lambda=1$ & {\cellcolor[HTML]{FEE1C4}} \color[HTML]{000000} -5.38 dB \\
 FL most helpful $\lambda=.75$ & {\cellcolor[HTML]{FEE0C1}} \color[HTML]{000000} -5.23 dB \\
 FL most harmful $\lambda=.75$ & {\cellcolor[HTML]{FEDEBD}} \color[HTML]{000000} -4.95 dB \\
 DIVINE most helpful & {\cellcolor[HTML]{FDDBB8}} \color[HTML]{000000} -4.62 dB \\
 FL most harmful $\lambda=1$ & {\cellcolor[HTML]{FDD8B2}} \color[HTML]{000000} -4.26 dB \\
 FL most helpful $\lambda=1$ & {\cellcolor[HTML]{FDD7B1}} \color[HTML]{000000} -4.13 dB \\
 DIVINE most inf. & {\cellcolor[HTML]{FDD6AE}} \color[HTML]{000000} -4.03 dB \\
 DIVINE most harmful & {\cellcolor[HTML]{FDD1A4}} \color[HTML]{000000} -3.34 dB \\
random & {\cellcolor[HTML]{A43503}} \color[HTML]{F1F1F1} 10.95 dB \\
 DIVINE least inf. & {\cellcolor[HTML]{7F2704}} \color[HTML]{F1F1F1} 13.66 dB \\
least inf. & {\cellcolor[HTML]{7F2704}} \color[HTML]{F1F1F1} 13.66 dB \\
 FL least inf. $\lambda=.75$ & {\cellcolor[HTML]{7F2704}} \color[HTML]{F1F1F1} 13.67 dB \\
 FL least inf. $\lambda=.5$ & {\cellcolor[HTML]{7F2704}} \color[HTML]{F1F1F1} 13.67 dB \\
 FL least inf. $\lambda=.25$ & {\cellcolor[HTML]{7F2704}} \color[HTML]{F1F1F1} 13.67 dB \\
 FL least inf. $\lambda=1$ & {\cellcolor[HTML]{7F2704}} \color[HTML]{F1F1F1} 13.67 dB \\
\bottomrule
\end{tabular}

    \caption*{DataInfEstimator}
\end{minipage}
\hfill
\begin{minipage}[t]{0.32\textwidth}
    \vspace{0pt}
    \centering
    \begin{tabular}{l|l|l}
\toprule
\midrule
most inf. & {\cellcolor[HTML]{FFF5EB}} \color[HTML]{000000} -8.28 dB \\
most harmful & {\cellcolor[HTML]{FFF3E6}} \color[HTML]{000000} -7.80 dB \\
most helpful & {\cellcolor[HTML]{FFF2E5}} \color[HTML]{000000} -7.67 dB \\
 FL most inf. $\lambda=.25$ & {\cellcolor[HTML]{FEEDDB}} \color[HTML]{000000} -6.73 dB \\
 AIDE & {\cellcolor[HTML]{FEECD9}} \color[HTML]{000000} -6.56 dB \\
 FL most inf. $\lambda=.5$ & {\cellcolor[HTML]{FEEBD7}} \color[HTML]{000000} -6.37 dB \\
 FL most harmful $\lambda=.25$ & {\cellcolor[HTML]{FEEBD7}} \color[HTML]{000000} -6.32 dB \\
 FL most helpful $\lambda=.25$ & {\cellcolor[HTML]{FEEAD5}} \color[HTML]{000000} -6.19 dB \\
 FL most inf. $\lambda=.75$ & {\cellcolor[HTML]{FEE8D2}} \color[HTML]{000000} -5.95 dB \\
 FL most helpful $\lambda=.5$ & {\cellcolor[HTML]{FEE8D2}} \color[HTML]{000000} -5.88 dB \\
 FL most harmful $\lambda=.5$ & {\cellcolor[HTML]{FEE7D1}} \color[HTML]{000000} -5.74 dB \\
 DIVINE most inf. & {\cellcolor[HTML]{FEE7D1}} \color[HTML]{000000} -5.72 dB \\
 FL most harmful $\lambda=.75$ & {\cellcolor[HTML]{FEE6CF}} \color[HTML]{000000} -5.54 dB \\
 DIVINE most harmful & {\cellcolor[HTML]{FEE6CE}} \color[HTML]{000000} -5.45 dB \\
 FL most inf. $\lambda=1$ & {\cellcolor[HTML]{FEE5CC}} \color[HTML]{000000} -5.43 dB \\
 FL most helpful $\lambda=.75$ & {\cellcolor[HTML]{FEE5CC}} \color[HTML]{000000} -5.38 dB \\
 FL most harmful $\lambda=1$ & {\cellcolor[HTML]{FEE2C6}} \color[HTML]{000000} -4.98 dB \\
 DIVINE most helpful & {\cellcolor[HTML]{FEE0C3}} \color[HTML]{000000} -4.85 dB \\
 FL most helpful $\lambda=1$ & {\cellcolor[HTML]{FEDCBB}} \color[HTML]{000000} -4.29 dB \\
random & {\cellcolor[HTML]{A53603}} \color[HTML]{F1F1F1} 10.95 dB \\
 DIVINE least inf. & {\cellcolor[HTML]{7F2704}} \color[HTML]{F1F1F1} 13.66 dB \\
least inf. & {\cellcolor[HTML]{7F2704}} \color[HTML]{F1F1F1} 13.66 dB \\
 FL least inf. $\lambda=1$ & {\cellcolor[HTML]{7F2704}} \color[HTML]{F1F1F1} 13.67 dB \\
 FL least inf. $\lambda=.75$ & {\cellcolor[HTML]{7F2704}} \color[HTML]{F1F1F1} 13.67 dB \\
 FL least inf. $\lambda=.5$ & {\cellcolor[HTML]{7F2704}} \color[HTML]{F1F1F1} 13.67 dB \\
 FL least inf. $\lambda=.25$ & {\cellcolor[HTML]{7F2704}} \color[HTML]{F1F1F1} 13.67 dB \\
\bottomrule
\end{tabular}

    \caption*{LESSEstimator}

\end{minipage}
\hfill
\begin{minipage}[t]{0.32\textwidth}
    \vspace{0pt}
    \centering
    \begin{tabular}{l|l|l}
\toprule
\midrule
random & {\cellcolor[HTML]{FFF5EB}} \color[HTML]{000000} 10.95 dB \\
 FL least inf. $\lambda=1$ & {\cellcolor[HTML]{FFF4E8}} \color[HTML]{000000} 11.37 dB \\
 DIVINE least inf. & {\cellcolor[HTML]{FFF3E6}} \color[HTML]{000000} 11.62 dB \\
 FL least inf. $\lambda=.75$ & {\cellcolor[HTML]{FFF0E1}} \color[HTML]{000000} 12.42 dB \\
 FL least inf. $\lambda=.5$ & {\cellcolor[HTML]{FFEFDF}} \color[HTML]{000000} 12.68 dB \\
 FL least inf. $\lambda=.25$ & {\cellcolor[HTML]{FFEEDD}} \color[HTML]{000000} 12.82 dB \\
least inf. & {\cellcolor[HTML]{FFEEDD}} \color[HTML]{000000} 12.88 dB \\
 FL most inf. $\lambda=1$ & {\cellcolor[HTML]{F67925}} \color[HTML]{F1F1F1} 29.03 dB \\
most inf. & {\cellcolor[HTML]{AD3803}} \color[HTML]{F1F1F1} 38.16 dB \\
 AIDE & {\cellcolor[HTML]{AD3803}} \color[HTML]{F1F1F1} 38.20 dB \\
 DIVINE most inf. & {\cellcolor[HTML]{993103}} \color[HTML]{F1F1F1} 40.10 dB \\
 FL most inf. $\lambda=.25$ & {\cellcolor[HTML]{812804}} \color[HTML]{F1F1F1} 42.36 dB \\
 FL most inf. $\lambda=.75$ & {\cellcolor[HTML]{7F2704}} \color[HTML]{F1F1F1} 42.69 dB \\
 FL most inf. $\lambda=.5$ & {\cellcolor[HTML]{7F2704}} \color[HTML]{F1F1F1} 42.73 dB \\
\bottomrule
\end{tabular}

    \caption*{BM25Estimator}
\end{minipage}
\caption{\textbf{Aggregate results.} auc $\sr$ $k={1,5,10,25}$. Per model results in Table \ref{tab:auc_per_model}. }\label{tab:auc}
\end{table*}
\FloatBarrier

\clearpage
\section{Case Study}\label{appendix:case_study}
\begin{figure*}[ht]  
\centering
  \includegraphics[width=1\textwidth, trim=0 5cm 0 0, clip]{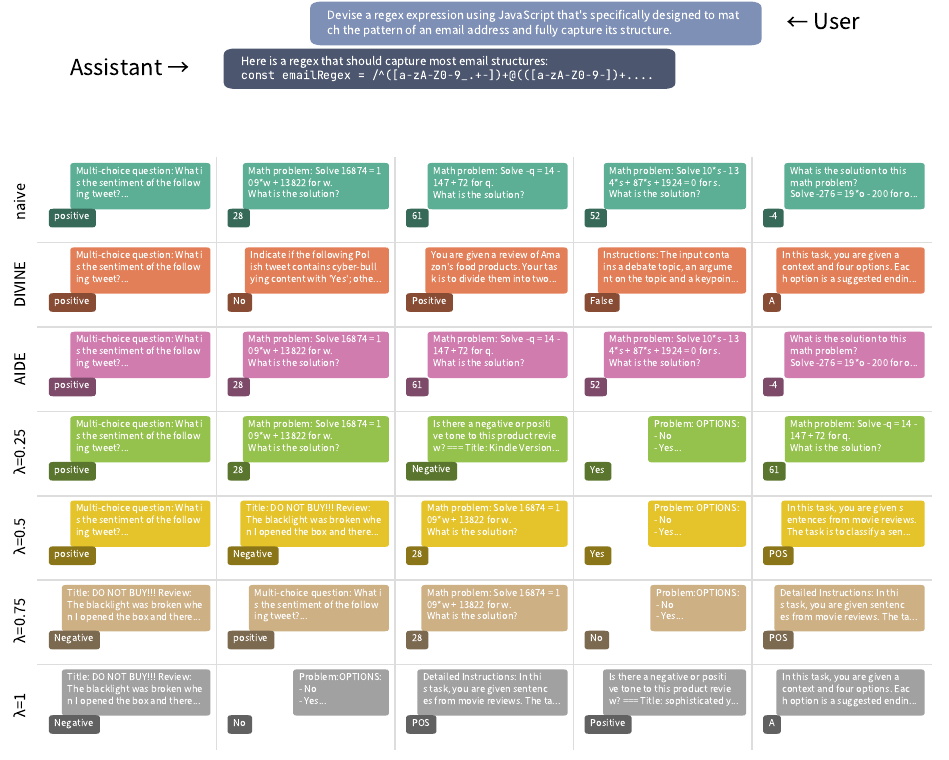}
  \caption {\textbf{Case Study}. Full selection for the test instance in Figure \ref{fig:onecol_case_study} ($k$=5, DataInf, most inf.).}\label{fig:full_case_study}
\end{figure*}
\FloatBarrier

\clearpage

\section{Per-Model Results}\label{appendix:per_model_results}
\begin{center}
\scriptsize
\setlength{\tabcolsep}{0.5pt}

\begin{minipage}[t]{0.32\textwidth} 
    \vspace{0pt}
    \centering
    \begin{tabular}{l|l|l}
\toprule
\midrule
Olmo2-1B & most inf. & {\cellcolor[HTML]{FFF7FB}} \color[HTML]{000000} -28.72 dB \\
Olmo2-1B & most harmful & {\cellcolor[HTML]{F0EAF4}} \color[HTML]{000000} -25.82 dB \\
Olmo2-1B & most helpful & {\cellcolor[HTML]{E9E5F1}} \color[HTML]{000000} -24.76 dB \\
Qwen2.5-0.5B & most inf. & {\cellcolor[HTML]{E7E3F0}} \color[HTML]{000000} -24.48 dB \\
Olmo2-1B &  FL most inf. $\lambda=.25$ & {\cellcolor[HTML]{E0DEED}} \color[HTML]{000000} -23.64 dB \\
Qwen2.5-0.5B & most harmful & {\cellcolor[HTML]{E0DEED}} \color[HTML]{000000} -23.62 dB \\
Qwen2.5-0.5B & most helpful & {\cellcolor[HTML]{E0DDED}} \color[HTML]{000000} -23.56 dB \\
Olmo2-1B &  FL most inf. $\lambda=.5$ & {\cellcolor[HTML]{DDDBEC}} \color[HTML]{000000} -23.21 dB \\
Olmo2-1B &  AIDE & {\cellcolor[HTML]{DCDAEB}} \color[HTML]{000000} -23.12 dB \\
Olmo2-1B &  FL most inf. $\lambda=.75$ & {\cellcolor[HTML]{DAD9EA}} \color[HTML]{000000} -22.84 dB \\
Olmo2-1B &  FL most harmful $\lambda=.25$ & {\cellcolor[HTML]{D7D6E9}} \color[HTML]{000000} -22.41 dB \\
Olmo2-1B &  FL most inf. $\lambda=1$ & {\cellcolor[HTML]{D5D5E8}} \color[HTML]{000000} -22.22 dB \\
Olmo2-1B &  FL most harmful $\lambda=.5$ & {\cellcolor[HTML]{D4D4E8}} \color[HTML]{000000} -22.06 dB \\
Olmo2-1B &  FL most helpful $\lambda=.25$ & {\cellcolor[HTML]{D3D4E7}} \color[HTML]{000000} -22.01 dB \\
Olmo2-1B &  FL most helpful $\lambda=.5$ & {\cellcolor[HTML]{D2D3E7}} \color[HTML]{000000} -21.81 dB \\
Olmo2-1B &  FL most harmful $\lambda=.75$ & {\cellcolor[HTML]{D1D2E6}} \color[HTML]{000000} -21.62 dB \\
Olmo2-1B &  FL most helpful $\lambda=.75$ & {\cellcolor[HTML]{CED0E6}} \color[HTML]{000000} -21.43 dB \\
Qwen2.5-0.5B &  FL most inf. $\lambda=.25$ & {\cellcolor[HTML]{CCCFE5}} \color[HTML]{000000} -21.19 dB \\
Olmo2-1B &  DIVINE most helpful & {\cellcolor[HTML]{CCCFE5}} \color[HTML]{000000} -21.14 dB \\
Qwen2.5-0.5B &  DIVINE most inf. & {\cellcolor[HTML]{CACEE5}} \color[HTML]{000000} -21.11 dB \\
Qwen2.5-0.5B &  DIVINE most helpful & {\cellcolor[HTML]{C8CDE4}} \color[HTML]{000000} -20.87 dB \\
Qwen2.5-0.5B &  AIDE & {\cellcolor[HTML]{C6CCE3}} \color[HTML]{000000} -20.73 dB \\
Qwen2.5-0.5B &  FL most helpful $\lambda=.25$ & {\cellcolor[HTML]{C5CCE3}} \color[HTML]{000000} -20.62 dB \\
Qwen2.5-0.5B &  FL most inf. $\lambda=.5$ & {\cellcolor[HTML]{C4CBE3}} \color[HTML]{000000} -20.55 dB \\
Qwen2.5-0.5B &  FL most harmful $\lambda=.25$ & {\cellcolor[HTML]{C4CBE3}} \color[HTML]{000000} -20.54 dB \\
Qwen2.5-0.5B &  FL most helpful $\lambda=.5$ & {\cellcolor[HTML]{C2CBE2}} \color[HTML]{000000} -20.44 dB \\
Olmo2-1B &  FL most harmful $\lambda=1$ & {\cellcolor[HTML]{C1CAE2}} \color[HTML]{000000} -20.32 dB \\
Qwen2.5-0.5B &  FL most inf. $\lambda=.75$ & {\cellcolor[HTML]{C0C9E2}} \color[HTML]{000000} -20.20 dB \\
Qwen2.5-0.5B &  FL most helpful $\lambda=.75$ & {\cellcolor[HTML]{C0C9E2}} \color[HTML]{000000} -20.20 dB \\
Qwen2.5-0.5B &  DIVINE most harmful & {\cellcolor[HTML]{C0C9E2}} \color[HTML]{000000} -20.11 dB \\
Qwen2.5-0.5B &  FL most harmful $\lambda=.5$ & {\cellcolor[HTML]{BFC9E1}} \color[HTML]{000000} -20.06 dB \\
Llama-3.2-1B & most harmful & {\cellcolor[HTML]{BFC9E1}} \color[HTML]{000000} -20.04 dB \\
Olmo2-1B &  FL most helpful $\lambda=1$ & {\cellcolor[HTML]{BCC7E1}} \color[HTML]{000000} -19.82 dB \\
Qwen2.5-0.5B &  FL most harmful $\lambda=.75$ & {\cellcolor[HTML]{BCC7E1}} \color[HTML]{000000} -19.80 dB \\
Qwen2.5-0.5B &  FL most inf. $\lambda=1$ & {\cellcolor[HTML]{BBC7E0}} \color[HTML]{000000} -19.77 dB \\
Llama-3.2-1B & most inf. & {\cellcolor[HTML]{BBC7E0}} \color[HTML]{000000} -19.76 dB \\
Qwen2.5-0.5B &  FL most helpful $\lambda=1$ & {\cellcolor[HTML]{B9C6E0}} \color[HTML]{000000} -19.61 dB \\
Qwen2.5-0.5B &  FL most harmful $\lambda=1$ & {\cellcolor[HTML]{B4C4DF}} \color[HTML]{000000} -19.19 dB \\
Llama-3.2-1B &  AIDE & {\cellcolor[HTML]{B4C4DF}} \color[HTML]{000000} -19.14 dB \\
Llama-3.2-1B &  FL most harmful $\lambda=.25$ & {\cellcolor[HTML]{B0C2DE}} \color[HTML]{000000} -18.81 dB \\
Llama-3.2-1B & most helpful & {\cellcolor[HTML]{AFC1DD}} \color[HTML]{000000} -18.73 dB \\
Llama-3.2-1B &  FL most inf. $\lambda=.25$ & {\cellcolor[HTML]{ADC1DD}} \color[HTML]{000000} -18.58 dB \\
Llama-3.2-1B &  FL most harmful $\lambda=.5$ & {\cellcolor[HTML]{ACC0DD}} \color[HTML]{000000} -18.45 dB \\
Llama-3.2-1B &  FL most inf. $\lambda=.5$ & {\cellcolor[HTML]{ABBFDC}} \color[HTML]{000000} -18.34 dB \\
Llama-3.2-1B &  FL most harmful $\lambda=.75$ & {\cellcolor[HTML]{A8BEDC}} \color[HTML]{000000} -18.14 dB \\
Llama-3.2-1B &  FL most inf. $\lambda=.75$ & {\cellcolor[HTML]{A8BEDC}} \color[HTML]{000000} -18.14 dB \\
Llama-3.2-1B &  FL most inf. $\lambda=1$ & {\cellcolor[HTML]{A2BCDA}} \color[HTML]{000000} -17.74 dB \\
Llama-3.2-1B &  FL most helpful $\lambda=.25$ & {\cellcolor[HTML]{A2BCDA}} \color[HTML]{000000} -17.74 dB \\
Llama-3.2-1B &  FL most harmful $\lambda=1$ & {\cellcolor[HTML]{A1BBDA}} \color[HTML]{000000} -17.64 dB \\
Llama-3.2-1B &  FL most helpful $\lambda=.5$ & {\cellcolor[HTML]{9EBAD9}} \color[HTML]{000000} -17.41 dB \\
Llama-3.2-1B &  DIVINE most harmful & {\cellcolor[HTML]{9EBAD9}} \color[HTML]{000000} -17.36 dB \\
Llama-3.2-1B &  DIVINE most inf. & {\cellcolor[HTML]{99B8D8}} \color[HTML]{000000} -17.06 dB \\
Llama-3.2-1B &  FL most helpful $\lambda=.75$ & {\cellcolor[HTML]{99B8D8}} \color[HTML]{000000} -17.01 dB \\
Llama-3.2-1B &  FL most helpful $\lambda=1$ & {\cellcolor[HTML]{8EB3D5}} \color[HTML]{000000} -16.25 dB \\
Llama-3.2-1B &  DIVINE most helpful & {\cellcolor[HTML]{8BB2D4}} \color[HTML]{000000} -16.08 dB \\
Olmo2-1B &  DIVINE most inf. & {\cellcolor[HTML]{83AFD3}} \color[HTML]{F1F1F1} -15.48 dB \\
Olmo2-1B &  DIVINE most harmful & {\cellcolor[HTML]{7EADD1}} \color[HTML]{F1F1F1} -15.15 dB \\
Qwen2.5-0.5B & random & {\cellcolor[HTML]{045382}} \color[HTML]{F1F1F1} -2.99 dB \\
Llama-3.2-1B & random & {\cellcolor[HTML]{034D79}} \color[HTML]{F1F1F1} -2.39 dB \\
Olmo2-1B & random & {\cellcolor[HTML]{034D79}} \color[HTML]{F1F1F1} -2.37 dB \\
Olmo2-1B & least inf. & {\cellcolor[HTML]{023858}} \color[HTML]{F1F1F1} -0.19 dB \\
Olmo2-1B &  DIVINE least inf. & {\cellcolor[HTML]{023858}} \color[HTML]{F1F1F1} -0.19 dB \\
Olmo2-1B &  FL least inf. $\lambda=.5$ & {\cellcolor[HTML]{023858}} \color[HTML]{F1F1F1} -0.18 dB \\
Olmo2-1B &  FL least inf. $\lambda=1$ & {\cellcolor[HTML]{023858}} \color[HTML]{F1F1F1} -0.18 dB \\
Olmo2-1B &  FL least inf. $\lambda=.75$ & {\cellcolor[HTML]{023858}} \color[HTML]{F1F1F1} -0.18 dB \\
Olmo2-1B &  FL least inf. $\lambda=.25$ & {\cellcolor[HTML]{023858}} \color[HTML]{F1F1F1} -0.18 dB \\
Qwen2.5-0.5B &  DIVINE least inf. & {\cellcolor[HTML]{023858}} \color[HTML]{F1F1F1} -0.15 dB \\
Qwen2.5-0.5B & least inf. & {\cellcolor[HTML]{023858}} \color[HTML]{F1F1F1} -0.14 dB \\
Qwen2.5-0.5B &  FL least inf. $\lambda=1$ & {\cellcolor[HTML]{023858}} \color[HTML]{F1F1F1} -0.13 dB \\
Qwen2.5-0.5B &  FL least inf. $\lambda=.75$ & {\cellcolor[HTML]{023858}} \color[HTML]{F1F1F1} -0.13 dB \\
Qwen2.5-0.5B &  FL least inf. $\lambda=.25$ & {\cellcolor[HTML]{023858}} \color[HTML]{F1F1F1} -0.13 dB \\
Qwen2.5-0.5B &  FL least inf. $\lambda=.5$ & {\cellcolor[HTML]{023858}} \color[HTML]{F1F1F1} -0.13 dB \\
Llama-3.2-1B &  DIVINE least inf. & {\cellcolor[HTML]{023858}} \color[HTML]{F1F1F1} -0.10 dB \\
Llama-3.2-1B & least inf. & {\cellcolor[HTML]{023858}} \color[HTML]{F1F1F1} -0.10 dB \\
Llama-3.2-1B &  FL least inf. $\lambda=1$ & {\cellcolor[HTML]{023858}} \color[HTML]{F1F1F1} -0.08 dB \\
Llama-3.2-1B &  FL least inf. $\lambda=.75$ & {\cellcolor[HTML]{023858}} \color[HTML]{F1F1F1} -0.08 dB \\
Llama-3.2-1B &  FL least inf. $\lambda=.5$ & {\cellcolor[HTML]{023858}} \color[HTML]{F1F1F1} -0.08 dB \\
Llama-3.2-1B &  FL least inf. $\lambda=.25$ & {\cellcolor[HTML]{023858}} \color[HTML]{F1F1F1} -0.08 dB \\
\bottomrule
\end{tabular}

    \captionof*{table}{DataInfEstimator}
\end{minipage}
\hfill
\begin{minipage}[t]{0.32\textwidth}
    \vspace{0pt}
    \centering
    \begin{tabular}{l|l|l}
\toprule
\midrule
Olmo2-1B & most inf. & {\cellcolor[HTML]{FFF7FB}} \color[HTML]{000000} -25.95 dB \\
Olmo2-1B & most harmful & {\cellcolor[HTML]{FAF2F8}} \color[HTML]{000000} -24.99 dB \\
Olmo2-1B & most helpful & {\cellcolor[HTML]{F4EEF6}} \color[HTML]{000000} -24.08 dB \\
Qwen2.5-0.5B & most inf. & {\cellcolor[HTML]{F3EDF5}} \color[HTML]{000000} -23.91 dB \\
Qwen2.5-0.5B & most helpful & {\cellcolor[HTML]{F1EBF5}} \color[HTML]{000000} -23.53 dB \\
Qwen2.5-0.5B & most harmful & {\cellcolor[HTML]{EDE8F3}} \color[HTML]{000000} -22.82 dB \\
Olmo2-1B &  FL most inf. $\lambda=.25$ & {\cellcolor[HTML]{ECE7F2}} \color[HTML]{000000} -22.70 dB \\
Olmo2-1B &  FL most inf. $\lambda=.5$ & {\cellcolor[HTML]{EBE6F2}} \color[HTML]{000000} -22.61 dB \\
Olmo2-1B &  FL most inf. $\lambda=.75$ & {\cellcolor[HTML]{E9E5F1}} \color[HTML]{000000} -22.39 dB \\
Olmo2-1B &  AIDE & {\cellcolor[HTML]{E9E5F1}} \color[HTML]{000000} -22.37 dB \\
Olmo2-1B &  DIVINE most inf. & {\cellcolor[HTML]{E8E4F0}} \color[HTML]{000000} -22.24 dB \\
Olmo2-1B &  FL most harmful $\lambda=.25$ & {\cellcolor[HTML]{E7E3F0}} \color[HTML]{000000} -22.08 dB \\
Olmo2-1B &  FL most inf. $\lambda=1$ & {\cellcolor[HTML]{E6E2EF}} \color[HTML]{000000} -21.91 dB \\
Olmo2-1B &  FL most harmful $\lambda=.5$ & {\cellcolor[HTML]{E5E1EF}} \color[HTML]{000000} -21.85 dB \\
Olmo2-1B &  DIVINE most harmful & {\cellcolor[HTML]{E3E0EE}} \color[HTML]{000000} -21.65 dB \\
Olmo2-1B &  FL most helpful $\lambda=.25$ & {\cellcolor[HTML]{E3E0EE}} \color[HTML]{000000} -21.62 dB \\
Olmo2-1B &  FL most harmful $\lambda=.75$ & {\cellcolor[HTML]{E2DFEE}} \color[HTML]{000000} -21.56 dB \\
Qwen2.5-0.5B &  FL most inf. $\lambda=.25$ & {\cellcolor[HTML]{E2DFEE}} \color[HTML]{000000} -21.54 dB \\
Olmo2-1B &  FL most helpful $\lambda=.5$ & {\cellcolor[HTML]{E0DDED}} \color[HTML]{000000} -21.29 dB \\
Qwen2.5-0.5B &  FL most inf. $\lambda=.5$ & {\cellcolor[HTML]{E0DDED}} \color[HTML]{000000} -21.29 dB \\
Qwen2.5-0.5B &  AIDE & {\cellcolor[HTML]{E0DDED}} \color[HTML]{000000} -21.21 dB \\
Qwen2.5-0.5B &  FL most helpful $\lambda=.25$ & {\cellcolor[HTML]{DFDDEC}} \color[HTML]{000000} -21.18 dB \\
Qwen2.5-0.5B &  DIVINE most inf. & {\cellcolor[HTML]{DEDCEC}} \color[HTML]{000000} -21.00 dB \\
Qwen2.5-0.5B &  FL most helpful $\lambda=.5$ & {\cellcolor[HTML]{DCDAEB}} \color[HTML]{000000} -20.85 dB \\
Olmo2-1B &  FL most helpful $\lambda=.75$ & {\cellcolor[HTML]{DCDAEB}} \color[HTML]{000000} -20.83 dB \\
Qwen2.5-0.5B &  FL most harmful $\lambda=.25$ & {\cellcolor[HTML]{DCDAEB}} \color[HTML]{000000} -20.83 dB \\
Olmo2-1B &  FL most harmful $\lambda=1$ & {\cellcolor[HTML]{DAD9EA}} \color[HTML]{000000} -20.60 dB \\
Olmo2-1B &  DIVINE most helpful & {\cellcolor[HTML]{D8D7E9}} \color[HTML]{000000} -20.39 dB \\
Qwen2.5-0.5B &  DIVINE most harmful & {\cellcolor[HTML]{D8D7E9}} \color[HTML]{000000} -20.38 dB \\
Qwen2.5-0.5B &  DIVINE most helpful & {\cellcolor[HTML]{D6D6E9}} \color[HTML]{000000} -20.18 dB \\
Olmo2-1B &  FL most helpful $\lambda=1$ & {\cellcolor[HTML]{D5D5E8}} \color[HTML]{000000} -20.05 dB \\
Qwen2.5-0.5B &  FL most helpful $\lambda=.75$ & {\cellcolor[HTML]{D4D4E8}} \color[HTML]{000000} -19.97 dB \\
Llama-3.2-1B & most inf. & {\cellcolor[HTML]{D1D2E6}} \color[HTML]{000000} -19.53 dB \\
Llama-3.2-1B & most helpful & {\cellcolor[HTML]{D0D1E6}} \color[HTML]{000000} -19.42 dB \\
Llama-3.2-1B & most harmful & {\cellcolor[HTML]{D0D1E6}} \color[HTML]{000000} -19.41 dB \\
Qwen2.5-0.5B &  FL most inf. $\lambda=.75$ & {\cellcolor[HTML]{CED0E6}} \color[HTML]{000000} -19.34 dB \\
Qwen2.5-0.5B &  FL most harmful $\lambda=1$ & {\cellcolor[HTML]{CED0E6}} \color[HTML]{000000} -19.29 dB \\
Qwen2.5-0.5B &  FL most helpful $\lambda=1$ & {\cellcolor[HTML]{CDD0E5}} \color[HTML]{000000} -19.27 dB \\
Qwen2.5-0.5B &  FL most inf. $\lambda=1$ & {\cellcolor[HTML]{CACEE5}} \color[HTML]{000000} -19.05 dB \\
Qwen2.5-0.5B &  FL most harmful $\lambda=.5$ & {\cellcolor[HTML]{CACEE5}} \color[HTML]{000000} -19.02 dB \\
Qwen2.5-0.5B &  FL most harmful $\lambda=.75$ & {\cellcolor[HTML]{C9CEE4}} \color[HTML]{000000} -18.90 dB \\
Llama-3.2-1B &  FL most inf. $\lambda=.25$ & {\cellcolor[HTML]{C4CBE3}} \color[HTML]{000000} -18.55 dB \\
Llama-3.2-1B &  AIDE & {\cellcolor[HTML]{C4CBE3}} \color[HTML]{000000} -18.53 dB \\
Llama-3.2-1B &  FL most harmful $\lambda=.25$ & {\cellcolor[HTML]{C2CBE2}} \color[HTML]{000000} -18.40 dB \\
Llama-3.2-1B &  FL most helpful $\lambda=.25$ & {\cellcolor[HTML]{C1CAE2}} \color[HTML]{000000} -18.29 dB \\
Llama-3.2-1B &  FL most inf. $\lambda=.75$ & {\cellcolor[HTML]{C0C9E2}} \color[HTML]{000000} -18.18 dB \\
Llama-3.2-1B &  FL most inf. $\lambda=.5$ & {\cellcolor[HTML]{BDC8E1}} \color[HTML]{000000} -18.05 dB \\
Llama-3.2-1B &  FL most harmful $\lambda=.75$ & {\cellcolor[HTML]{BDC8E1}} \color[HTML]{000000} -18.01 dB \\
Llama-3.2-1B &  FL most helpful $\lambda=.5$ & {\cellcolor[HTML]{BDC8E1}} \color[HTML]{000000} -17.97 dB \\
Llama-3.2-1B &  FL most harmful $\lambda=.5$ & {\cellcolor[HTML]{BCC7E1}} \color[HTML]{000000} -17.92 dB \\
Llama-3.2-1B &  FL most inf. $\lambda=1$ & {\cellcolor[HTML]{BBC7E0}} \color[HTML]{000000} -17.81 dB \\
Llama-3.2-1B &  FL most harmful $\lambda=1$ & {\cellcolor[HTML]{BBC7E0}} \color[HTML]{000000} -17.78 dB \\
Llama-3.2-1B &  FL most helpful $\lambda=.75$ & {\cellcolor[HTML]{B7C5DF}} \color[HTML]{000000} -17.55 dB \\
Llama-3.2-1B &  DIVINE most harmful & {\cellcolor[HTML]{B4C4DF}} \color[HTML]{000000} -17.31 dB \\
Llama-3.2-1B &  DIVINE most inf. & {\cellcolor[HTML]{B4C4DF}} \color[HTML]{000000} -17.26 dB \\
Llama-3.2-1B &  DIVINE most helpful & {\cellcolor[HTML]{ADC1DD}} \color[HTML]{000000} -16.82 dB \\
Llama-3.2-1B &  FL most helpful $\lambda=1$ & {\cellcolor[HTML]{ACC0DD}} \color[HTML]{000000} -16.68 dB \\
Qwen2.5-0.5B & random & {\cellcolor[HTML]{045687}} \color[HTML]{F1F1F1} -2.99 dB \\
Llama-3.2-1B & random & {\cellcolor[HTML]{034F7D}} \color[HTML]{F1F1F1} -2.39 dB \\
Olmo2-1B & random & {\cellcolor[HTML]{034F7D}} \color[HTML]{F1F1F1} -2.37 dB \\
Olmo2-1B & least inf. & {\cellcolor[HTML]{023858}} \color[HTML]{F1F1F1} -0.19 dB \\
Olmo2-1B &  DIVINE least inf. & {\cellcolor[HTML]{023858}} \color[HTML]{F1F1F1} -0.19 dB \\
Olmo2-1B &  FL least inf. $\lambda=.5$ & {\cellcolor[HTML]{023858}} \color[HTML]{F1F1F1} -0.18 dB \\
Olmo2-1B &  FL least inf. $\lambda=.25$ & {\cellcolor[HTML]{023858}} \color[HTML]{F1F1F1} -0.18 dB \\
Olmo2-1B &  FL least inf. $\lambda=1$ & {\cellcolor[HTML]{023858}} \color[HTML]{F1F1F1} -0.18 dB \\
Olmo2-1B &  FL least inf. $\lambda=.75$ & {\cellcolor[HTML]{023858}} \color[HTML]{F1F1F1} -0.18 dB \\
Qwen2.5-0.5B &  DIVINE least inf. & {\cellcolor[HTML]{023858}} \color[HTML]{F1F1F1} -0.14 dB \\
Qwen2.5-0.5B & least inf. & {\cellcolor[HTML]{023858}} \color[HTML]{F1F1F1} -0.14 dB \\
Qwen2.5-0.5B &  FL least inf. $\lambda=.75$ & {\cellcolor[HTML]{023858}} \color[HTML]{F1F1F1} -0.13 dB \\
Qwen2.5-0.5B &  FL least inf. $\lambda=1$ & {\cellcolor[HTML]{023858}} \color[HTML]{F1F1F1} -0.13 dB \\
Qwen2.5-0.5B &  FL least inf. $\lambda=.5$ & {\cellcolor[HTML]{023858}} \color[HTML]{F1F1F1} -0.13 dB \\
Qwen2.5-0.5B &  FL least inf. $\lambda=.25$ & {\cellcolor[HTML]{023858}} \color[HTML]{F1F1F1} -0.13 dB \\
Llama-3.2-1B &  DIVINE least inf. & {\cellcolor[HTML]{023858}} \color[HTML]{F1F1F1} -0.10 dB \\
Llama-3.2-1B & least inf. & {\cellcolor[HTML]{023858}} \color[HTML]{F1F1F1} -0.10 dB \\
Llama-3.2-1B &  FL least inf. $\lambda=.5$ & {\cellcolor[HTML]{023858}} \color[HTML]{F1F1F1} -0.09 dB \\
Llama-3.2-1B &  FL least inf. $\lambda=1$ & {\cellcolor[HTML]{023858}} \color[HTML]{F1F1F1} -0.09 dB \\
Llama-3.2-1B &  FL least inf. $\lambda=.75$ & {\cellcolor[HTML]{023858}} \color[HTML]{F1F1F1} -0.09 dB \\
Llama-3.2-1B &  FL least inf. $\lambda=.25$ & {\cellcolor[HTML]{023858}} \color[HTML]{F1F1F1} -0.09 dB \\
\bottomrule
\end{tabular}

    \captionof*{table}{LESSEstimator}
\end{minipage}
\hfill
\begin{minipage}[t]{0.32\textwidth}
    \vspace{0pt}
    \centering
    \begin{tabular}{l|l|l}
\toprule
\midrule
Qwen2.5-0.5B &  FL least inf. $\lambda=1$ & {\cellcolor[HTML]{FFF7FB}} \color[HTML]{000000} -3.50 dB \\
Qwen2.5-0.5B &  DIVINE least inf. & {\cellcolor[HTML]{FDF5FA}} \color[HTML]{000000} -3.06 dB \\
Qwen2.5-0.5B & random & {\cellcolor[HTML]{FDF5FA}} \color[HTML]{000000} -2.99 dB \\
Llama-3.2-1B & random & {\cellcolor[HTML]{FBF3F9}} \color[HTML]{000000} -2.39 dB \\
Olmo2-1B &  DIVINE least inf. & {\cellcolor[HTML]{FBF3F9}} \color[HTML]{000000} -2.38 dB \\
Olmo2-1B & random & {\cellcolor[HTML]{FBF3F9}} \color[HTML]{000000} -2.37 dB \\
Olmo2-1B &  FL least inf. $\lambda=1$ & {\cellcolor[HTML]{FBF3F9}} \color[HTML]{000000} -2.32 dB \\
Llama-3.2-1B &  DIVINE least inf. & {\cellcolor[HTML]{FAF2F8}} \color[HTML]{000000} -2.14 dB \\
Qwen2.5-0.5B &  FL least inf. $\lambda=.75$ & {\cellcolor[HTML]{F9F2F8}} \color[HTML]{000000} -1.96 dB \\
Olmo2-1B &  FL least inf. $\lambda=.75$ & {\cellcolor[HTML]{F8F1F8}} \color[HTML]{000000} -1.76 dB \\
Llama-3.2-1B &  FL least inf. $\lambda=1$ & {\cellcolor[HTML]{F8F1F8}} \color[HTML]{000000} -1.61 dB \\
Qwen2.5-0.5B &  FL least inf. $\lambda=.5$ & {\cellcolor[HTML]{F7F0F7}} \color[HTML]{000000} -1.58 dB \\
Qwen2.5-0.5B &  FL least inf. $\lambda=.25$ & {\cellcolor[HTML]{F7F0F7}} \color[HTML]{000000} -1.47 dB \\
Olmo2-1B &  FL least inf. $\lambda=.5$ & {\cellcolor[HTML]{F7F0F7}} \color[HTML]{000000} -1.44 dB \\
Qwen2.5-0.5B & least inf. & {\cellcolor[HTML]{F7F0F7}} \color[HTML]{000000} -1.43 dB \\
Olmo2-1B &  FL least inf. $\lambda=.25$ & {\cellcolor[HTML]{F6EFF7}} \color[HTML]{000000} -1.25 dB \\
Olmo2-1B & least inf. & {\cellcolor[HTML]{F5EFF6}} \color[HTML]{000000} -1.13 dB \\
Llama-3.2-1B &  FL least inf. $\lambda=.75$ & {\cellcolor[HTML]{F5EFF6}} \color[HTML]{000000} -1.03 dB \\
Llama-3.2-1B &  FL least inf. $\lambda=.5$ & {\cellcolor[HTML]{F4EEF6}} \color[HTML]{000000} -0.79 dB \\
Llama-3.2-1B &  FL least inf. $\lambda=.25$ & {\cellcolor[HTML]{F4EDF6}} \color[HTML]{000000} -0.60 dB \\
Llama-3.2-1B & least inf. & {\cellcolor[HTML]{F3EDF5}} \color[HTML]{000000} -0.52 dB \\
Olmo2-1B &  FL most inf. $\lambda=1$ & {\cellcolor[HTML]{8FB4D6}} \color[HTML]{000000} 12.78 dB \\
Qwen2.5-0.5B &  FL most inf. $\lambda=1$ & {\cellcolor[HTML]{71A8CE}} \color[HTML]{F1F1F1} 15.65 dB \\
Llama-3.2-1B &  FL most inf. $\lambda=1$ & {\cellcolor[HTML]{5EA0CA}} \color[HTML]{F1F1F1} 17.05 dB \\
Olmo2-1B &  DIVINE most inf. & {\cellcolor[HTML]{358FC0}} \color[HTML]{F1F1F1} 20.23 dB \\
Olmo2-1B &  FL most inf. $\lambda=.5$ & {\cellcolor[HTML]{2786BB}} \color[HTML]{F1F1F1} 21.46 dB \\
Olmo2-1B &  FL most inf. $\lambda=.25$ & {\cellcolor[HTML]{2786BB}} \color[HTML]{F1F1F1} 21.46 dB \\
Qwen2.5-0.5B &  AIDE & {\cellcolor[HTML]{2685BB}} \color[HTML]{F1F1F1} 21.62 dB \\
Olmo2-1B &  FL most inf. $\lambda=.75$ & {\cellcolor[HTML]{2685BB}} \color[HTML]{F1F1F1} 21.65 dB \\
Qwen2.5-0.5B & most inf. & {\cellcolor[HTML]{2685BB}} \color[HTML]{F1F1F1} 21.69 dB \\
Olmo2-1B &  AIDE & {\cellcolor[HTML]{2484BA}} \color[HTML]{F1F1F1} 21.86 dB \\
Olmo2-1B & most inf. & {\cellcolor[HTML]{2383BA}} \color[HTML]{F1F1F1} 21.91 dB \\
Llama-3.2-1B &  FL most inf. $\lambda=.75$ & {\cellcolor[HTML]{2383BA}} \color[HTML]{F1F1F1} 22.01 dB \\
Llama-3.2-1B &  FL most inf. $\lambda=.5$ & {\cellcolor[HTML]{1E80B8}} \color[HTML]{F1F1F1} 22.42 dB \\
Llama-3.2-1B &  DIVINE most inf. & {\cellcolor[HTML]{1C7FB8}} \color[HTML]{F1F1F1} 22.50 dB \\
Llama-3.2-1B &  AIDE & {\cellcolor[HTML]{1B7EB7}} \color[HTML]{F1F1F1} 22.68 dB \\
Llama-3.2-1B &  FL most inf. $\lambda=.25$ & {\cellcolor[HTML]{197DB7}} \color[HTML]{F1F1F1} 22.87 dB \\
Llama-3.2-1B & most inf. & {\cellcolor[HTML]{187CB6}} \color[HTML]{F1F1F1} 22.99 dB \\
Qwen2.5-0.5B &  DIVINE most inf. & {\cellcolor[HTML]{0569A4}} \color[HTML]{F1F1F1} 26.35 dB \\
Qwen2.5-0.5B &  FL most inf. $\lambda=.25$ & {\cellcolor[HTML]{023C5F}} \color[HTML]{F1F1F1} 33.63 dB \\
Qwen2.5-0.5B &  FL most inf. $\lambda=.5$ & {\cellcolor[HTML]{023858}} \color[HTML]{F1F1F1} 34.30 dB \\
Qwen2.5-0.5B &  FL most inf. $\lambda=.75$ & {\cellcolor[HTML]{023858}} \color[HTML]{F1F1F1} 34.30 dB \\
\bottomrule
\end{tabular}

    \captionof*{table}{BM25Estimator}
\end{minipage}
\captionsetup{hypcap=false}
\captionof{table}{\textbf{Per-model results. $\sr$ $k=10$.}} 
\end{center}

\begin{table*}[htbp]
\centering
\scriptsize
\setlength{\tabcolsep}{0.5pt} 
\begin{minipage}[t]{0.32\textwidth} 
    \vspace{0pt}
    \centering
    \begin{tabular}{l|l|l}
\toprule
\midrule
Olmo2-1B & most inf. & {\cellcolor[HTML]{FFF5EB}} \color[HTML]{000000} -14.13 dB \\
Olmo2-1B & most harmful & {\cellcolor[HTML]{FEE9D4}} \color[HTML]{000000} -11.32 dB \\
Olmo2-1B & most helpful & {\cellcolor[HTML]{FEE6CE}} \color[HTML]{000000} -10.57 dB \\
Qwen2.5-0.5B & most inf. & {\cellcolor[HTML]{FEE5CC}} \color[HTML]{000000} -10.48 dB \\
Olmo2-1B &  AIDE & {\cellcolor[HTML]{FEE1C4}} \color[HTML]{000000} -9.87 dB \\
Qwen2.5-0.5B & most helpful & {\cellcolor[HTML]{FEDEBF}} \color[HTML]{000000} -9.42 dB \\
Qwen2.5-0.5B & most harmful & {\cellcolor[HTML]{FEDEBF}} \color[HTML]{000000} -9.40 dB \\
Olmo2-1B &  FL most inf. $\lambda=.25$ & {\cellcolor[HTML]{FEDCB9}} \color[HTML]{000000} -8.92 dB \\
Olmo2-1B &  FL most inf. $\lambda=.5$ & {\cellcolor[HTML]{FDD9B4}} \color[HTML]{000000} -8.54 dB \\
Qwen2.5-0.5B &  FL most inf. $\lambda=.25$ & {\cellcolor[HTML]{FDD5AD}} \color[HTML]{000000} -7.97 dB \\
Olmo2-1B &  FL most harmful $\lambda=.25$ & {\cellcolor[HTML]{FDD4AA}} \color[HTML]{000000} -7.76 dB \\
Olmo2-1B &  FL most inf. $\lambda=.75$ & {\cellcolor[HTML]{FDD4AA}} \color[HTML]{000000} -7.76 dB \\
Qwen2.5-0.5B &  AIDE & {\cellcolor[HTML]{FDD2A6}} \color[HTML]{000000} -7.48 dB \\
Qwen2.5-0.5B &  FL most inf. $\lambda=.5$ & {\cellcolor[HTML]{FDD2A6}} \color[HTML]{000000} -7.46 dB \\
Olmo2-1B &  FL most helpful $\lambda=.25$ & {\cellcolor[HTML]{FDD2A6}} \color[HTML]{000000} -7.46 dB \\
Olmo2-1B &  FL most harmful $\lambda=.5$ & {\cellcolor[HTML]{FDD2A6}} \color[HTML]{000000} -7.41 dB \\
Qwen2.5-0.5B &  FL most helpful $\lambda=.25$ & {\cellcolor[HTML]{FDD1A3}} \color[HTML]{000000} -7.28 dB \\
Qwen2.5-0.5B &  DIVINE most inf. & {\cellcolor[HTML]{FDD1A3}} \color[HTML]{000000} -7.24 dB \\
Qwen2.5-0.5B &  FL most harmful $\lambda=.25$ & {\cellcolor[HTML]{FDD1A3}} \color[HTML]{000000} -7.22 dB \\
Olmo2-1B &  FL most helpful $\lambda=.5$ & {\cellcolor[HTML]{FDD1A3}} \color[HTML]{000000} -7.21 dB \\
Qwen2.5-0.5B &  FL most helpful $\lambda=.5$ & {\cellcolor[HTML]{FDCFA0}} \color[HTML]{000000} -7.03 dB \\
Qwen2.5-0.5B &  FL most inf. $\lambda=.75$ & {\cellcolor[HTML]{FDCFA0}} \color[HTML]{000000} -7.02 dB \\
Olmo2-1B &  FL most helpful $\lambda=.75$ & {\cellcolor[HTML]{FDCD9C}} \color[HTML]{000000} -6.81 dB \\
Olmo2-1B &  DIVINE most helpful & {\cellcolor[HTML]{FDCD9C}} \color[HTML]{000000} -6.73 dB \\
Qwen2.5-0.5B &  DIVINE most helpful & {\cellcolor[HTML]{FDCB9B}} \color[HTML]{000000} -6.66 dB \\
Qwen2.5-0.5B &  FL most helpful $\lambda=.75$ & {\cellcolor[HTML]{FDCB9B}} \color[HTML]{000000} -6.65 dB \\
Qwen2.5-0.5B &  FL most harmful $\lambda=.5$ & {\cellcolor[HTML]{FDCA99}} \color[HTML]{000000} -6.57 dB \\
Qwen2.5-0.5B &  FL most inf. $\lambda=1$ & {\cellcolor[HTML]{FDC895}} \color[HTML]{000000} -6.39 dB \\
Olmo2-1B &  FL most inf. $\lambda=1$ & {\cellcolor[HTML]{FDC895}} \color[HTML]{000000} -6.37 dB \\
Qwen2.5-0.5B &  FL most harmful $\lambda=.75$ & {\cellcolor[HTML]{FDC794}} \color[HTML]{000000} -6.20 dB \\
Qwen2.5-0.5B &  DIVINE most harmful & {\cellcolor[HTML]{FDC692}} \color[HTML]{000000} -6.10 dB \\
Llama-3.2-1B & most harmful & {\cellcolor[HTML]{FDC590}} \color[HTML]{000000} -5.98 dB \\
Llama-3.2-1B & most inf. & {\cellcolor[HTML]{FDC38D}} \color[HTML]{000000} -5.84 dB \\
Qwen2.5-0.5B &  FL most helpful $\lambda=1$ & {\cellcolor[HTML]{FDC38D}} \color[HTML]{000000} -5.79 dB \\
Qwen2.5-0.5B &  FL most harmful $\lambda=1$ & {\cellcolor[HTML]{FDC189}} \color[HTML]{000000} -5.55 dB \\
Olmo2-1B &  FL most helpful $\lambda=1$ & {\cellcolor[HTML]{FDBF86}} \color[HTML]{000000} -5.37 dB \\
Llama-3.2-1B &  FL most harmful $\lambda=.25$ & {\cellcolor[HTML]{FDBD83}} \color[HTML]{000000} -5.12 dB \\
Llama-3.2-1B &  AIDE & {\cellcolor[HTML]{FDBB81}} \color[HTML]{000000} -5.06 dB \\
Llama-3.2-1B &  FL most inf. $\lambda=.25$ & {\cellcolor[HTML]{FDBA7F}} \color[HTML]{000000} -4.99 dB \\
Llama-3.2-1B &  FL most inf. $\lambda=.5$ & {\cellcolor[HTML]{FDB87C}} \color[HTML]{000000} -4.76 dB \\
Llama-3.2-1B &  FL most harmful $\lambda=.5$ & {\cellcolor[HTML]{FDB87C}} \color[HTML]{000000} -4.76 dB \\
Llama-3.2-1B & most helpful & {\cellcolor[HTML]{FDB87C}} \color[HTML]{000000} -4.74 dB \\
Llama-3.2-1B &  FL most inf. $\lambda=.75$ & {\cellcolor[HTML]{FDB678}} \color[HTML]{000000} -4.48 dB \\
Olmo2-1B &  FL most harmful $\lambda=.75$ & {\cellcolor[HTML]{FDB678}} \color[HTML]{000000} -4.47 dB \\
Llama-3.2-1B &  FL most harmful $\lambda=.75$ & {\cellcolor[HTML]{FDB576}} \color[HTML]{000000} -4.41 dB \\
Llama-3.2-1B &  FL most helpful $\lambda=.25$ & {\cellcolor[HTML]{FDB170}} \color[HTML]{000000} -3.93 dB \\
Llama-3.2-1B &  FL most inf. $\lambda=1$ & {\cellcolor[HTML]{FDB06E}} \color[HTML]{000000} -3.88 dB \\
Llama-3.2-1B &  FL most harmful $\lambda=1$ & {\cellcolor[HTML]{FDAF6C}} \color[HTML]{000000} -3.77 dB \\
Olmo2-1B &  FL most harmful $\lambda=1$ & {\cellcolor[HTML]{FDAE6A}} \color[HTML]{000000} -3.69 dB \\
Llama-3.2-1B &  FL most helpful $\lambda=.5$ & {\cellcolor[HTML]{FDAE6A}} \color[HTML]{000000} -3.66 dB \\
Llama-3.2-1B &  DIVINE most harmful & {\cellcolor[HTML]{FDAE6A}} \color[HTML]{000000} -3.58 dB \\
Llama-3.2-1B &  DIVINE most inf. & {\cellcolor[HTML]{FDAB66}} \color[HTML]{000000} -3.35 dB \\
Llama-3.2-1B &  FL most helpful $\lambda=.75$ & {\cellcolor[HTML]{FDA965}} \color[HTML]{000000} -3.23 dB \\
Olmo2-1B &  DIVINE most inf. & {\cellcolor[HTML]{FDA55F}} \color[HTML]{000000} -2.71 dB \\
Llama-3.2-1B &  FL most helpful $\lambda=1$ & {\cellcolor[HTML]{FDA057}} \color[HTML]{000000} -2.18 dB \\
Llama-3.2-1B &  DIVINE most helpful & {\cellcolor[HTML]{FD9F56}} \color[HTML]{000000} -2.16 dB \\
Olmo2-1B &  DIVINE most harmful & {\cellcolor[HTML]{FD994D}} \color[HTML]{000000} -1.50 dB \\
Qwen2.5-0.5B & random & {\cellcolor[HTML]{A13403}} \color[HTML]{F1F1F1} 10.61 dB \\
Llama-3.2-1B & random & {\cellcolor[HTML]{9C3203}} \color[HTML]{F1F1F1} 11.09 dB \\
Olmo2-1B & random & {\cellcolor[HTML]{9B3203}} \color[HTML]{F1F1F1} 11.12 dB \\
Olmo2-1B & least inf. & {\cellcolor[HTML]{7F2704}} \color[HTML]{F1F1F1} 13.61 dB \\
Olmo2-1B &  DIVINE least inf. & {\cellcolor[HTML]{7F2704}} \color[HTML]{F1F1F1} 13.61 dB \\
Olmo2-1B &  FL least inf. $\lambda=1$ & {\cellcolor[HTML]{7F2704}} \color[HTML]{F1F1F1} 13.63 dB \\
Olmo2-1B &  FL least inf. $\lambda=.25$ & {\cellcolor[HTML]{7F2704}} \color[HTML]{F1F1F1} 13.63 dB \\
Olmo2-1B &  FL least inf. $\lambda=.5$ & {\cellcolor[HTML]{7F2704}} \color[HTML]{F1F1F1} 13.63 dB \\
Olmo2-1B &  FL least inf. $\lambda=.75$ & {\cellcolor[HTML]{7F2704}} \color[HTML]{F1F1F1} 13.63 dB \\
Qwen2.5-0.5B &  DIVINE least inf. & {\cellcolor[HTML]{7F2704}} \color[HTML]{F1F1F1} 13.66 dB \\
Qwen2.5-0.5B & least inf. & {\cellcolor[HTML]{7F2704}} \color[HTML]{F1F1F1} 13.66 dB \\
Qwen2.5-0.5B &  FL least inf. $\lambda=.75$ & {\cellcolor[HTML]{7F2704}} \color[HTML]{F1F1F1} 13.67 dB \\
Qwen2.5-0.5B &  FL least inf. $\lambda=.5$ & {\cellcolor[HTML]{7F2704}} \color[HTML]{F1F1F1} 13.67 dB \\
Qwen2.5-0.5B &  FL least inf. $\lambda=.25$ & {\cellcolor[HTML]{7F2704}} \color[HTML]{F1F1F1} 13.67 dB \\
Qwen2.5-0.5B &  FL least inf. $\lambda=1$ & {\cellcolor[HTML]{7F2704}} \color[HTML]{F1F1F1} 13.67 dB \\
Llama-3.2-1B &  DIVINE least inf. & {\cellcolor[HTML]{7F2704}} \color[HTML]{F1F1F1} 13.71 dB \\
Llama-3.2-1B & least inf. & {\cellcolor[HTML]{7F2704}} \color[HTML]{F1F1F1} 13.71 dB \\
Llama-3.2-1B &  FL least inf. $\lambda=1$ & {\cellcolor[HTML]{7F2704}} \color[HTML]{F1F1F1} 13.72 dB \\
Llama-3.2-1B &  FL least inf. $\lambda=.75$ & {\cellcolor[HTML]{7F2704}} \color[HTML]{F1F1F1} 13.72 dB \\
Llama-3.2-1B &  FL least inf. $\lambda=.5$ & {\cellcolor[HTML]{7F2704}} \color[HTML]{F1F1F1} 13.72 dB \\
Llama-3.2-1B &  FL least inf. $\lambda=.25$ & {\cellcolor[HTML]{7F2704}} \color[HTML]{F1F1F1} 13.72 dB \\
\bottomrule
\end{tabular}

    \caption*{DataInfEstimator}
\end{minipage}
\hfill
\begin{minipage}[t]{0.32\textwidth}
    \vspace{0pt}
    \centering
    \begin{tabular}{l|l|l}
\toprule
\midrule
Olmo2-1B & most inf. & {\cellcolor[HTML]{FFF5EB}} \color[HTML]{000000} -11.86 dB \\
Olmo2-1B & most harmful & {\cellcolor[HTML]{FFF0E2}} \color[HTML]{000000} -10.82 dB \\
Qwen2.5-0.5B & most inf. & {\cellcolor[HTML]{FEECD9}} \color[HTML]{000000} -9.86 dB \\
Olmo2-1B & most helpful & {\cellcolor[HTML]{FEECD9}} \color[HTML]{000000} -9.78 dB \\
Qwen2.5-0.5B & most helpful & {\cellcolor[HTML]{FEE9D4}} \color[HTML]{000000} -9.33 dB \\
Olmo2-1B &  AIDE & {\cellcolor[HTML]{FEE7D1}} \color[HTML]{000000} -8.87 dB \\
Qwen2.5-0.5B & most harmful & {\cellcolor[HTML]{FEE6CF}} \color[HTML]{000000} -8.72 dB \\
Olmo2-1B &  FL most inf. $\lambda=.25$ & {\cellcolor[HTML]{FEE3C8}} \color[HTML]{000000} -8.23 dB \\
Olmo2-1B &  DIVINE most inf. & {\cellcolor[HTML]{FEE2C7}} \color[HTML]{000000} -8.16 dB \\
Olmo2-1B &  FL most inf. $\lambda=.5$ & {\cellcolor[HTML]{FEE2C6}} \color[HTML]{000000} -8.05 dB \\
Qwen2.5-0.5B &  FL most inf. $\lambda=.25$ & {\cellcolor[HTML]{FEE1C4}} \color[HTML]{000000} -7.96 dB \\
Olmo2-1B &  FL most inf. $\lambda=.75$ & {\cellcolor[HTML]{FEE0C3}} \color[HTML]{000000} -7.83 dB \\
Qwen2.5-0.5B &  FL most helpful $\lambda=.25$ & {\cellcolor[HTML]{FEDEBF}} \color[HTML]{000000} -7.55 dB \\
Olmo2-1B &  FL most harmful $\lambda=.25$ & {\cellcolor[HTML]{FEDEBF}} \color[HTML]{000000} -7.54 dB \\
Qwen2.5-0.5B &  AIDE & {\cellcolor[HTML]{FEDEBF}} \color[HTML]{000000} -7.48 dB \\
Qwen2.5-0.5B &  FL most inf. $\lambda=.5$ & {\cellcolor[HTML]{FEDEBD}} \color[HTML]{000000} -7.37 dB \\
Olmo2-1B &  DIVINE most harmful & {\cellcolor[HTML]{FEDDBC}} \color[HTML]{000000} -7.33 dB \\
Olmo2-1B &  FL most harmful $\lambda=.5$ & {\cellcolor[HTML]{FEDDBC}} \color[HTML]{000000} -7.30 dB \\
Qwen2.5-0.5B &  FL most helpful $\lambda=.5$ & {\cellcolor[HTML]{FEDCBB}} \color[HTML]{000000} -7.21 dB \\
Olmo2-1B &  FL most inf. $\lambda=1$ & {\cellcolor[HTML]{FEDCBB}} \color[HTML]{000000} -7.21 dB \\
Qwen2.5-0.5B &  FL most harmful $\lambda=.25$ & {\cellcolor[HTML]{FEDCBB}} \color[HTML]{000000} -7.17 dB \\
Olmo2-1B &  FL most helpful $\lambda=.25$ & {\cellcolor[HTML]{FEDCB9}} \color[HTML]{000000} -7.12 dB \\
Qwen2.5-0.5B &  DIVINE most inf. & {\cellcolor[HTML]{FEDCB9}} \color[HTML]{000000} -7.09 dB \\
Olmo2-1B &  FL most harmful $\lambda=.75$ & {\cellcolor[HTML]{FDDBB8}} \color[HTML]{000000} -7.06 dB \\
Olmo2-1B &  FL most helpful $\lambda=.5$ & {\cellcolor[HTML]{FDDAB6}} \color[HTML]{000000} -6.87 dB \\
Qwen2.5-0.5B &  FL most helpful $\lambda=.75$ & {\cellcolor[HTML]{FDD8B2}} \color[HTML]{000000} -6.62 dB \\
Olmo2-1B &  FL most helpful $\lambda=.75$ & {\cellcolor[HTML]{FDD7AF}} \color[HTML]{000000} -6.45 dB \\
Qwen2.5-0.5B &  DIVINE most harmful & {\cellcolor[HTML]{FDD7AF}} \color[HTML]{000000} -6.38 dB \\
Olmo2-1B &  DIVINE most helpful & {\cellcolor[HTML]{FDD6AE}} \color[HTML]{000000} -6.32 dB \\
Qwen2.5-0.5B &  DIVINE most helpful & {\cellcolor[HTML]{FDD5AD}} \color[HTML]{000000} -6.23 dB \\
Qwen2.5-0.5B &  FL most inf. $\lambda=.75$ & {\cellcolor[HTML]{FDD5AD}} \color[HTML]{000000} -6.19 dB \\
Olmo2-1B &  FL most harmful $\lambda=1$ & {\cellcolor[HTML]{FDD3A9}} \color[HTML]{000000} -5.97 dB \\
Qwen2.5-0.5B &  FL most harmful $\lambda=.5$ & {\cellcolor[HTML]{FDD3A9}} \color[HTML]{000000} -5.95 dB \\
Qwen2.5-0.5B &  FL most inf. $\lambda=1$ & {\cellcolor[HTML]{FDD3A9}} \color[HTML]{000000} -5.89 dB \\
Qwen2.5-0.5B &  FL most harmful $\lambda=.75$ & {\cellcolor[HTML]{FDD1A4}} \color[HTML]{000000} -5.58 dB \\
Llama-3.2-1B & most inf. & {\cellcolor[HTML]{FDD1A4}} \color[HTML]{000000} -5.57 dB \\
Qwen2.5-0.5B &  FL most helpful $\lambda=1$ & {\cellcolor[HTML]{FDD1A3}} \color[HTML]{000000} -5.54 dB \\
Llama-3.2-1B & most harmful & {\cellcolor[HTML]{FDD1A3}} \color[HTML]{000000} -5.51 dB \\
Qwen2.5-0.5B &  FL most harmful $\lambda=1$ & {\cellcolor[HTML]{FDD0A2}} \color[HTML]{000000} -5.46 dB \\
Olmo2-1B &  FL most helpful $\lambda=1$ & {\cellcolor[HTML]{FDD0A2}} \color[HTML]{000000} -5.42 dB \\
Llama-3.2-1B & most helpful & {\cellcolor[HTML]{FDCFA0}} \color[HTML]{000000} -5.37 dB \\
Llama-3.2-1B &  FL most inf. $\lambda=.25$ & {\cellcolor[HTML]{FDC997}} \color[HTML]{000000} -4.86 dB \\
Llama-3.2-1B &  FL most harmful $\lambda=.25$ & {\cellcolor[HTML]{FDC997}} \color[HTML]{000000} -4.80 dB \\
Llama-3.2-1B &  FL most helpful $\lambda=.25$ & {\cellcolor[HTML]{FDC692}} \color[HTML]{000000} -4.55 dB \\
Llama-3.2-1B &  FL most inf. $\lambda=.5$ & {\cellcolor[HTML]{FDC692}} \color[HTML]{000000} -4.54 dB \\
Llama-3.2-1B &  AIDE & {\cellcolor[HTML]{FDC692}} \color[HTML]{000000} -4.52 dB \\
Llama-3.2-1B &  FL most inf. $\lambda=.75$ & {\cellcolor[HTML]{FDC692}} \color[HTML]{000000} -4.47 dB \\
Llama-3.2-1B &  FL most harmful $\lambda=.5$ & {\cellcolor[HTML]{FDC590}} \color[HTML]{000000} -4.44 dB \\
Llama-3.2-1B &  FL most harmful $\lambda=.75$ & {\cellcolor[HTML]{FDC590}} \color[HTML]{000000} -4.38 dB \\
Llama-3.2-1B &  FL most helpful $\lambda=.5$ & {\cellcolor[HTML]{FDC38D}} \color[HTML]{000000} -4.22 dB \\
Llama-3.2-1B &  FL most inf. $\lambda=1$ & {\cellcolor[HTML]{FDBF86}} \color[HTML]{000000} -3.87 dB \\
Llama-3.2-1B &  FL most harmful $\lambda=1$ & {\cellcolor[HTML]{FDBF86}} \color[HTML]{000000} -3.81 dB \\
Llama-3.2-1B &  FL most helpful $\lambda=.75$ & {\cellcolor[HTML]{FDBE84}} \color[HTML]{000000} -3.72 dB \\
Llama-3.2-1B &  DIVINE most harmful & {\cellcolor[HTML]{FDBB81}} \color[HTML]{000000} -3.57 dB \\
Llama-3.2-1B &  DIVINE most inf. & {\cellcolor[HTML]{FDBA7F}} \color[HTML]{000000} -3.41 dB \\
Llama-3.2-1B &  DIVINE most helpful & {\cellcolor[HTML]{FDB576}} \color[HTML]{000000} -2.92 dB \\
Llama-3.2-1B &  FL most helpful $\lambda=1$ & {\cellcolor[HTML]{FDB271}} \color[HTML]{000000} -2.59 dB \\
Qwen2.5-0.5B & random & {\cellcolor[HTML]{A53603}} \color[HTML]{F1F1F1} 10.61 dB \\
Llama-3.2-1B & random & {\cellcolor[HTML]{9F3303}} \color[HTML]{F1F1F1} 11.09 dB \\
Olmo2-1B & random & {\cellcolor[HTML]{9E3303}} \color[HTML]{F1F1F1} 11.12 dB \\
Olmo2-1B & least inf. & {\cellcolor[HTML]{7F2704}} \color[HTML]{F1F1F1} 13.61 dB \\
Olmo2-1B &  DIVINE least inf. & {\cellcolor[HTML]{7F2704}} \color[HTML]{F1F1F1} 13.61 dB \\
Olmo2-1B &  FL least inf. $\lambda=.75$ & {\cellcolor[HTML]{7F2704}} \color[HTML]{F1F1F1} 13.63 dB \\
Olmo2-1B &  FL least inf. $\lambda=1$ & {\cellcolor[HTML]{7F2704}} \color[HTML]{F1F1F1} 13.63 dB \\
Olmo2-1B &  FL least inf. $\lambda=.5$ & {\cellcolor[HTML]{7F2704}} \color[HTML]{F1F1F1} 13.63 dB \\
Olmo2-1B &  FL least inf. $\lambda=.25$ & {\cellcolor[HTML]{7F2704}} \color[HTML]{F1F1F1} 13.63 dB \\
Qwen2.5-0.5B &  DIVINE least inf. & {\cellcolor[HTML]{7F2704}} \color[HTML]{F1F1F1} 13.66 dB \\
Qwen2.5-0.5B & least inf. & {\cellcolor[HTML]{7F2704}} \color[HTML]{F1F1F1} 13.66 dB \\
Qwen2.5-0.5B &  FL least inf. $\lambda=.5$ & {\cellcolor[HTML]{7F2704}} \color[HTML]{F1F1F1} 13.67 dB \\
Qwen2.5-0.5B &  FL least inf. $\lambda=.25$ & {\cellcolor[HTML]{7F2704}} \color[HTML]{F1F1F1} 13.67 dB \\
Qwen2.5-0.5B &  FL least inf. $\lambda=1$ & {\cellcolor[HTML]{7F2704}} \color[HTML]{F1F1F1} 13.67 dB \\
Qwen2.5-0.5B &  FL least inf. $\lambda=.75$ & {\cellcolor[HTML]{7F2704}} \color[HTML]{F1F1F1} 13.67 dB \\
Llama-3.2-1B &  DIVINE least inf. & {\cellcolor[HTML]{7F2704}} \color[HTML]{F1F1F1} 13.70 dB \\
Llama-3.2-1B & least inf. & {\cellcolor[HTML]{7F2704}} \color[HTML]{F1F1F1} 13.70 dB \\
Llama-3.2-1B &  FL least inf. $\lambda=.5$ & {\cellcolor[HTML]{7F2704}} \color[HTML]{F1F1F1} 13.71 dB \\
Llama-3.2-1B &  FL least inf. $\lambda=.25$ & {\cellcolor[HTML]{7F2704}} \color[HTML]{F1F1F1} 13.71 dB \\
Llama-3.2-1B &  FL least inf. $\lambda=.75$ & {\cellcolor[HTML]{7F2704}} \color[HTML]{F1F1F1} 13.71 dB \\
Llama-3.2-1B &  FL least inf. $\lambda=1$ & {\cellcolor[HTML]{7F2704}} \color[HTML]{F1F1F1} 13.71 dB \\
\bottomrule
\end{tabular}

    \caption*{LESSEstimator}

\end{minipage}
\hfill
\begin{minipage}[t]{0.32\textwidth}
    \vspace{0pt}
    \centering
    \begin{tabular}{l|l|l}
\toprule
\midrule
Qwen2.5-0.5B &  FL least inf. $\lambda=1$ & {\cellcolor[HTML]{FFF5EB}} \color[HTML]{000000} 10.52 dB \\
Qwen2.5-0.5B & random & {\cellcolor[HTML]{FFF5EB}} \color[HTML]{000000} 10.61 dB \\
Qwen2.5-0.5B &  DIVINE least inf. & {\cellcolor[HTML]{FFF4E8}} \color[HTML]{000000} 11.02 dB \\
Llama-3.2-1B & random & {\cellcolor[HTML]{FFF3E7}} \color[HTML]{000000} 11.09 dB \\
Olmo2-1B & random & {\cellcolor[HTML]{FFF3E7}} \color[HTML]{000000} 11.12 dB \\
Olmo2-1B &  FL least inf. $\lambda=1$ & {\cellcolor[HTML]{FFF3E6}} \color[HTML]{000000} 11.31 dB \\
Olmo2-1B &  DIVINE least inf. & {\cellcolor[HTML]{FFF1E4}} \color[HTML]{000000} 11.75 dB \\
Llama-3.2-1B &  DIVINE least inf. & {\cellcolor[HTML]{FFF0E2}} \color[HTML]{000000} 12.02 dB \\
Qwen2.5-0.5B &  FL least inf. $\lambda=.75$ & {\cellcolor[HTML]{FFF0E2}} \color[HTML]{000000} 12.06 dB \\
Llama-3.2-1B &  FL least inf. $\lambda=1$ & {\cellcolor[HTML]{FFF0E1}} \color[HTML]{000000} 12.13 dB \\
Olmo2-1B &  FL least inf. $\lambda=.75$ & {\cellcolor[HTML]{FFEFE0}} \color[HTML]{000000} 12.26 dB \\
Qwen2.5-0.5B &  FL least inf. $\lambda=.5$ & {\cellcolor[HTML]{FFEFDF}} \color[HTML]{000000} 12.41 dB \\
Olmo2-1B &  FL least inf. $\lambda=.5$ & {\cellcolor[HTML]{FFEEDE}} \color[HTML]{000000} 12.52 dB \\
Qwen2.5-0.5B &  FL least inf. $\lambda=.25$ & {\cellcolor[HTML]{FFEEDE}} \color[HTML]{000000} 12.52 dB \\
Qwen2.5-0.5B & least inf. & {\cellcolor[HTML]{FFEEDE}} \color[HTML]{000000} 12.54 dB \\
Olmo2-1B &  FL least inf. $\lambda=.25$ & {\cellcolor[HTML]{FFEEDD}} \color[HTML]{000000} 12.66 dB \\
Olmo2-1B & least inf. & {\cellcolor[HTML]{FFEEDD}} \color[HTML]{000000} 12.74 dB \\
Llama-3.2-1B &  FL least inf. $\lambda=.75$ & {\cellcolor[HTML]{FEEDDC}} \color[HTML]{000000} 12.89 dB \\
Llama-3.2-1B &  FL least inf. $\lambda=.5$ & {\cellcolor[HTML]{FEEDDB}} \color[HTML]{000000} 13.09 dB \\
Llama-3.2-1B &  FL least inf. $\lambda=.25$ & {\cellcolor[HTML]{FEECDA}} \color[HTML]{000000} 13.24 dB \\
Llama-3.2-1B & least inf. & {\cellcolor[HTML]{FEECDA}} \color[HTML]{000000} 13.31 dB \\
Olmo2-1B &  FL most inf. $\lambda=1$ & {\cellcolor[HTML]{FD984B}} \color[HTML]{000000} 27.14 dB \\
Qwen2.5-0.5B &  FL most inf. $\lambda=1$ & {\cellcolor[HTML]{FD8E3D}} \color[HTML]{F1F1F1} 28.52 dB \\
Llama-3.2-1B &  FL most inf. $\lambda=1$ & {\cellcolor[HTML]{F87D29}} \color[HTML]{F1F1F1} 30.69 dB \\
Olmo2-1B &  DIVINE most inf. & {\cellcolor[HTML]{EE6410}} \color[HTML]{F1F1F1} 33.73 dB \\
Olmo2-1B &  FL most inf. $\lambda=.5$ & {\cellcolor[HTML]{E85D0C}} \color[HTML]{F1F1F1} 34.69 dB \\
Olmo2-1B &  FL most inf. $\lambda=.25$ & {\cellcolor[HTML]{E85D0C}} \color[HTML]{F1F1F1} 34.79 dB \\
Olmo2-1B &  FL most inf. $\lambda=.75$ & {\cellcolor[HTML]{E75B0B}} \color[HTML]{F1F1F1} 34.96 dB \\
Olmo2-1B & most inf. & {\cellcolor[HTML]{E75B0B}} \color[HTML]{F1F1F1} 35.08 dB \\
Olmo2-1B &  AIDE & {\cellcolor[HTML]{E5590A}} \color[HTML]{F1F1F1} 35.38 dB \\
Llama-3.2-1B &  FL most inf. $\lambda=.75$ & {\cellcolor[HTML]{D64701}} \color[HTML]{F1F1F1} 37.86 dB \\
Llama-3.2-1B &  AIDE & {\cellcolor[HTML]{D54601}} \color[HTML]{F1F1F1} 38.03 dB \\
Llama-3.2-1B &  FL most inf. $\lambda=.25$ & {\cellcolor[HTML]{D34601}} \color[HTML]{F1F1F1} 38.10 dB \\
Llama-3.2-1B & most inf. & {\cellcolor[HTML]{D34601}} \color[HTML]{F1F1F1} 38.11 dB \\
Llama-3.2-1B &  FL most inf. $\lambda=.5$ & {\cellcolor[HTML]{D04501}} \color[HTML]{F1F1F1} 38.43 dB \\
Llama-3.2-1B &  DIVINE most inf. & {\cellcolor[HTML]{BE3F02}} \color[HTML]{F1F1F1} 39.91 dB \\
Qwen2.5-0.5B & most inf. & {\cellcolor[HTML]{BE3F02}} \color[HTML]{F1F1F1} 39.97 dB \\
Qwen2.5-0.5B &  AIDE & {\cellcolor[HTML]{BE3F02}} \color[HTML]{F1F1F1} 40.00 dB \\
Qwen2.5-0.5B &  DIVINE most inf. & {\cellcolor[HTML]{A13403}} \color[HTML]{F1F1F1} 42.68 dB \\
Qwen2.5-0.5B &  FL most inf. $\lambda=.25$ & {\cellcolor[HTML]{812804}} \color[HTML]{F1F1F1} 46.26 dB \\
Qwen2.5-0.5B &  FL most inf. $\lambda=.5$ & {\cellcolor[HTML]{7F2704}} \color[HTML]{F1F1F1} 46.66 dB \\
Qwen2.5-0.5B &  FL most inf. $\lambda=.75$ & {\cellcolor[HTML]{7F2704}} \color[HTML]{F1F1F1} 46.68 dB \\
\bottomrule
\end{tabular}

    \caption*{BM25Estimator}
\end{minipage}
\caption{\textbf{Per-model results.} auc $\sr$ $k={1,5,10,25}$.}\label{tab:auc_per_model}
\end{table*}
\FloatBarrier

\clearpage

\FloatBarrier
\twocolumn
\section{Model Fine-tuning}\label{appendix:finetuning}
We fine-tune all base models using LoRA for one epoch on the full \textit{Tülu3} \hfcite{https://huggingface.co/datasets/allenai/tulu-3-sft-olmo-2-mixture-0225}{lambert_tulu_2025} instruction-fine tuning dataset. We report performance on the \verb|OLMES| evaluation suite \cite{gu_olmes_2025} in Table \ref{tab:benchmark_results}.
\begin{table}[htbp]
  \centering 
  \small
  \begin{tabular}{|l|l|}
  \hline
  Precision & bfloat16 \\
  Optimizer & AdamW (torch fused) \\
  Learning rate & $1\times10^{-4}$ \\
  LR scheduler & Linear \\
  Weight decay & 0.0 \\
  Max grad norm & 1.0 \\
  \hline
  LoRA rank ($r$) & 16 \\
  LoRA alpha & 32 \\
  LoRA dropout & 0.1 \\
  LoRA bias & none \\
  Target modules &
  \makecell[l]{%
    Qwen \& Llama:\\
    \hspace{1em}\texttt{q\_proj, k\_proj,}\\
    \hspace{1em}\texttt{ v\_proj, o\_proj}\\
    Olmo2:\\
    \hspace{1em}\texttt{q\_proj, c\_attn,}\\
    \hspace{1em}\texttt{ v\_proj}
  } \\

  Trainable params & LoRA only \\
  \hline
  Train batch size / device & 4 \\
  Gradient accumulation & 8 \\
  Effective batch size & 32 \\
  Training epochs & 1 \\
  Max sequence length & 1024 \\
  Gradient checkpointing & False \\
  \hline
  Seed & 42 \\
  \hline
  \end{tabular}
  \caption{LoRA fine-tuning hyperparameters}
  \label{tab:LoRA_hparams}
\end{table}

\begin{minipage}{\textwidth}
  \centering
  \resizebox{1\textwidth}{!}{
  \setlength{\tabcolsep}{0.0pt} 
  \begin{tabular}{l|c|cccccccccccccccccccc}
\toprule
Task & \rotatebox{45}{Avg} & \rotatebox{45}{AGIEval} & \rotatebox{45}{ARC\_C} & \rotatebox{45}{ARC\_E} & \rotatebox{45}{BBH} & \rotatebox{45}{BoolQ} & \rotatebox{45}{CSQA} & \rotatebox{45}{CoQA} & \rotatebox{45}{DROP} & \rotatebox{45}{GSM8K} & \rotatebox{45}{HSwag} & \rotatebox{45}{JPRDY} & \rotatebox{45}{MMLU} & \rotatebox{45}{MMLU-Pro} & \rotatebox{45}{NatQs} & \rotatebox{45}{OBQA} & \rotatebox{45}{PIQA} & \rotatebox{45}{SIQA} & \rotatebox{45}{SQuAD} & \rotatebox{45}{TriviaQA} & \rotatebox{45}{WinoG} \\
 &  &  &  &  &  &  &  &  &  &  &  &  &  &  &  &  &  &  &  &  &  \\
\midrule
\textbf{Fine-tuned Models} &  &  &  &  &  &  &  &  &  &  &  &  &  &  &  &  &  &  &  &  &  \\
Llama-3.2-1B & .44 & .24 & .38 & .60 & .32 & .67 & .50 & .65 & .25 & .08 & .53 & .53 & .30 & .16 & .16 & .39 & .67 & .47 & .73 & .48 & .58 \\
OLMo-2-0425-1B & .52 & .34 & .47 & .74 & .30 & .69 & .60 & .69 & \textbf{.35} & .36 & .60 & \textbf{.63} & .43 & .19 & .19 & .51 & .71 & .56 & .80 & \textbf{.55} & \textbf{.61} \\
Qwen2.5-0.5B & .46 & .37 & .49 & .71 & .32 & .66 & .58 & .60 & .24 & .34 & .48 & .32 & \textbf{.49} & .23 & .12 & .54 & .66 & .56 & .73 & .19 & .54 \\
\hline
\textbf{External Models} &  &  &  &  &  &  &  &  &  &  &  &  &  &  &  &  &  &  &  &  &  \\
Llama-3.2-1B-Instruct & .51 & .35 & .49 & .73 & \textbf{.37} & .64 & .63 & .67 & .29 & .35 & .54 & .51 & .46 & \textbf{.25} & \textbf{.20} & \textbf{.54} & .71 & .57 & .79 & .47 & .59 \\
OLMo-2-0425-1B-SFT & \textbf{.53} & .36 & .48 & \textbf{.75} & .32 & \textbf{.75} & \textbf{.65} & \textbf{.72} & .33 & \textbf{.43} & \textbf{.61} & .57 & .44 & .21 & .17 & .51 & \textbf{.72} & \textbf{.58} & \textbf{.83} & .48 & .60 \\
Qwen2.5-0.5B-Instruct & .44 & \textbf{.38} & \textbf{.49} & .71 & .31 & .68 & .60 & .41 & .25 & .30 & .49 & .33 & .48 & .23 & .09 & .54 & .67 & .56 & .74 & .09 & .53 \\
\bottomrule
\end{tabular}
  }
  \vspace{1em}
  \captionsetup{justification=centering, singlelinecheck=false, skip=0pt,hypcap=false}
  \captionof{table}{\textbf{Benchmark results.} Performance on the OLMES evaluation suite. Best model in bold. Note that \textit{OLMo-2-0425-1B-SFT} is trained for one additional epoch and without LoRA.}
  \label{tab:benchmark_results}
\end{minipage}

\end{document}